\documentclass[sigconf,screen]{acmart}
\renewcommand\footnotetextcopyrightpermission[1]{}
\AtBeginDocument{%
  }

\acmConference[Preprint]{}{Feb 25}{2026}

\usepackage{enumitem}
\usepackage{pifont}
\usepackage{algorithmic}
\usepackage{tikz}
\usepackage[utf8]{inputenc}

\usepackage{amssymb}
\usepackage[linesnumbered,ruled,vlined]{algorithm2e}
\usepackage[switch]{lineno}
\usepackage{xspace}
\usepackage{booktabs}
\usepackage{comment}

\usepackage{textcomp}
\usepackage{array}
\usepackage{wrapfig}
\usepackage{soul}
\usepackage[dvipsnames]{xcolor}

\usepackage{hyperref}
\hypersetup{
    colorlinks=true,  
    linkcolor=blue,   
    citecolor=red,    
    filecolor=cyan,   
    urlcolor=magenta  
}

\definecolor{ForestGreen}{RGB}{34,139,34}
\definecolor{amber}{rgb}{1.0, 0.49, 0.0}

\usepackage[labelfont=bf]{caption}
\usepackage[T1]{fontenc}
\usepackage{fancyvrb}

\def\ricky#1{{#1}}

\definecolor{azure}{rgb}{0.0, 0.5, 1.0}

\newcommand{\eg}{{\it e.g.}}

\newcommand{\ie}{{\it i.e.}}
\newcommand{\vs}{\textit{vs.}\xspace}

\newcommand*\circled[1]{\tikz[baseline=(char.base)]{\node[shape=circle,draw,inner sep=1.5pt] (char) {#1};}}

\SetKwSty{text}
\SetArgSty{texttt}
\SetDataSty{}
\SetKwInput{KwInput}{\textcolor{blue}{\textbf{Input}}}
\SetKwInput{KwOutput}{\textcolor{blue}{\textbf{Output}}}
\SetKwIF{If}{ElseIf}{Else}{\textcolor{blue}{\textbf{if}}}{\textcolor{blue}{\textbf{then}}}{\textcolor{blue}{\textbf{else if}}}{\textcolor{blue}{\textbf{else}}}{\textcolor{blue}{\textbf{endif}}}
\SetKwFor{For}{\textcolor{blue}{\textbf{for}}}{\textcolor{blue}{\textbf{do}}}{\textcolor{blue}
{\textbf{endfor}}}
\SetKwFunction{Length}{length}
\SetKwFunction{Max}{max}
\SetKwProg{Fn}{\textcolor{blue}{\textbf{function}}}{}{}
\SetKw{KwRet}{\textcolor{blue}{\textbf{return}}}
\SetKwComment{Comment}{/* }{ */}

\usepackage{acronym}
\usepackage{xspace}

\newcommand{\sys}{\texttt{RLHFless}\xspace}

\acrodef{ML}[ML]{Machine Learning}

\acrodef{NSF}[NSF]{National Science Foundation}

\acrodef{AI}[AI]{Artificial Intelligence}
\newcommand{\AI}{\ac{AI}\xspace}

\acrodef{FL}[FL]{Federated Learning}

\acrodef{CL}[CL]{Critical Learning}

\acrodef{AC}[AC]{Attacking-Critical}

\acrodef{CAGR}[CAGR]{compound annual growth rate}

\acrodef{CCT}[CCT]{Center for Computation and Technology}

\acrodef{SLO}[SLO]{service level objective}

\acrodefplural{SLOs}[SLO]{service level objectives}

\acrodef{RL}[RL]{reinforcement learning}
\newcommand{\RL}{\ac{RL}\xspace}

\acrodef{DRL}[DRL]{deep reinforcement learning}

\acrodef{VM}[VM]{virtual machine}

\acrodefplural{VM}[VMs]{virtual machines}

\acrodef{ITC}[ITC]{Innovation \& Technology Commercialization}

\acrodef{DAG}[DAG]{directed acyclic graph}

\acrodefplural{DAG}[DAGs]{directed acyclic graphs}

\acrodef{SFA}[SFA]{single point authentication}

\acrodef{HPC}[HPC]{high-performance computing}

\acrodef{SBIR}[SBIR]{Small Business Innovation Research}

\acrodef{IoT}[IoT]{Internet of Things}

\acrodef{DML}[DML]{distributed machine learning}

\acrodef{GNN}[GNN]{graph neural network}

\acrodefplural{GNN}[GNNs]{graph neural networks}

\acrodef{BSR}[BSR]{backdoor success rate}

\acrodef{BTA}[BTA]{backdoor task accuracy}

\acrodef{ATT}[ATT]{App Tracking Transparency}

\acrodef{DNN}[DNN]{deep neural network}

\acrodef{DNNs}[DNNs]{deep neural networks}

\acrodef{KL}[KL]{Kullback–Leibler}
\newcommand{\KL}{\ac{KL}\xspace}

\acrodef{IaaS}[IaaS]{Infrastructure-as-a-Service}

\acrodef{CaaS}[CaaS]{Container-as-a-Service}

\acrodef{TRPO}[TRPO]{Trust Region Policy Optimization}

\acrodef{CPO}[CPO]{Constrained Policy Optimization}

\acrodef{PPO}[PPO]{Proximal Policy Optimization}
\newcommand{\PPO}{\ac{PPO}\xspace}

\acrodef{TV}[TV]{Total Variation}

\acrodef{PAC}[PAC]{Probably Approximately
Correct}

\acrodef{ACI}[ACI]{Azure Container Instances}

\acrodef{NLP}[NLP]{Natural Language Processing}

\acrodef{GAE}[GAE]{Generalized Advantage Estimation}

\acrodef{IS}[IS]{Importance Sampling}

\acrodef{CV}[CV]{coefficient of variance}

\acrodefplural{CV}[CVs]{coefficient of variances}

\acrodef{LLM}[LLM]{Large Language Model}
\newcommand{\LLM}{\ac{LLM}\xspace}

\acrodef{GNS}[GNS]{gradient noise scale}

\acrodef{SOTA}[SOTA]{state-of-the-art}
\newcommand{\SOTA}{\ac{SOTA}\xspace}

\acrodef{SSP}[SSP]{Stale Synchronous Parallel}

\acrodef{CDF}[CDF]{cumulative distribution function}

\acrodef{PDF}[PDF]{probability density function}

\acrodef{RPC}[RPC]{remote procedure call}

\acrodef{MARL}[MARL]{multi-agent reinforcement learning}

\acrodef{SARL}[SARL]{single-agent reinforcement learning}

\acrodef{MDP}[MDP]{Markov Decision Process}

\acrodef{CTDE}[CTDE]{centralized training \& decentralized execution}

\acrodef{MAPD}[MAPD]{Multi-Agent Policy Distance}

\acrodef{IPPO}[IPPO]{Independent Proximal Policy Optimization}

\acrodef{MPE}[MPE]{Multi-Agent Particle Environment}

\acrodef{SMAC}[SMAC]{StarCraft Multi-Agent Challenge}

\acrodef{DDPG}[DDPG]{Deep Deterministic Policy Gradient}

\acrodef{DQN}[DQN]{Deep Q-Network}

\acrodef{MAPPO}[MAPPO]{Multi-Agent Proximal Policy Optimization}

\acrodef{MADDPG}[MADDPG]{Multi-Agent Deep Deterministic Policy Gradient}

\acrodef{IQL}[IQL]{Independent Q-Learning}

\acrodef{KDE}[KDE]{Kernel Density Estimation}

\acrodef{VDN}[VDN]{Value-Decomposition Networks}

\acrodef{SAC}[SAC]{Soft Actor-Critic}

\acrodef{QMIX}[QMIX]{Monotonic Value Function Factorisation}

\acrodef{IDDPG}[IDDPG]{Independent Deep Deterministic Policy Gradient}

\acrodef{RNN}[RNN]{recurrent neural network}
\newcommand{\RNN}{\ac{RNN}\xspace}

\acrodef{CNN}[CNN]{convolutional neural network}
\newcommand{\CNN}{\ac{CNN}\xspace}

\acrodef{MLP}[MLP]{multi-layer perception}
\newcommand{\MLP}{\ac{MLP}\xspace}

\acrodef{NN}[NN]{neural network}

\acrodef{PER}[PER]{Prioritized Experience Replay}

\acrodef{RLHF}[RLHF]{Reinforcement Learning from Human Feedback}
\newcommand{\RLHF}{\ac{RLHF}\xspace}

\acrodef{RLAIF}[RLAIF]{Reinforcement Learning from AI Feedback}

\acrodef{GRPO}[GRPO]{Group Relative Policy Optimization}
\newcommand{\GRPO}{\ac{GRPO}\xspace}

\acrodef{DPO}[DPO]{Direct Preference Optimization}
\newcommand{\DPO}{\ac{DPO}\xspace}

\acrodef{SFT}[SFT]{Supervised Fine-tuning}
\newcommand{\SFT}{\ac{SFT}\xspace}

\acrodef{FSDP}[FSDP]{Fully Sharded Data Parallel}
\newcommand{\FSDP}{\ac{FSDP}\xspace}

\acrodef{MoE}[MoE]{Mixture-of-Experts}

\acrodef{RM}[RM]{Reward Model}
\newcommand{\RM}{\ac{RM}\xspace}

\acrodef{OOM}[OOM]{out-of-memory}
\newcommand{\OOM}{\ac{OOM}\xspace}

\acrodef{KV}[KV]{key-value}
\newcommand{\KV}{\ac{KV}\xspace}

\acrodef{FLOPs}[FLOPs]{floating point operations}

\acrodef{TPOT}[TPOT]{time-per-output-token}
\newcommand{\TPOT}{\ac{TPOT}\xspace}

\acrodef{EWMA}[EWMA]{exponentially weighted moving average}
\newcommand{\EWMA}{\ac{EWMA}\xspace}

\acrodef{LoRA}[LoRA]{Low-Rank Adaption}

\acrodef{NCCL}[NCCL]{NVIDIA Collective Communications Library}
\newcommand{\NCCL}{\ac{NCCL}\xspace}

\acrodefplural{LLM}[LLMs]{Large Language Models}
\newcommand{\LLMs}{\acp{LLM}\xspace}

\begin{document}

\title{\sys: Serverless Computing for Efficient RLHF}

\author{Rui Wei}
\email{rwei7@stevens.edu}
\affiliation{
  \institution{Stevens Institute of Technology}
}

\author{Hanfei Yu}
\email{hyu42@stevens.edu}
\affiliation{
  \institution{Stevens Institute of Technology}
}

\author{Shubham Jain}
\email{sjain71@stevens.edu}
\affiliation{
  \institution{Stevens Institute of Technology}
}

\author{Yogarajan Sivakumar}
\email{ysivakum@stevens.edu}
\affiliation{
  \institution{Stevens Institute of Technology}
}

\author{Devesh Tiwari}
\email{d.tiwari@northeastern.edu}
\affiliation{
  \institution{Northeastern University}
}

\author{Jian Li}
\email{jian.li.3@stonybrook.edu}
\affiliation{
  \institution{Stony Brook University}
}

\author{Seung-Jong Park}
\email{seung-jong.park@mst.edu}
\affiliation{
  \institution{\mbox{Missouri University of Science \& Technology}}
}

\author{Hao Wang}
\email{hwang9@stevens.edu}
\affiliation{
  \institution{Stevens Institute of Technology}
}

\renewcommand{\shortauthors}{Wei et al.}

\begin{abstract}
Reinforcement Learning from Human Feedback (RLHF) has been widely applied to Large Language Model (LLM) post-training to align model outputs with human preferences. Recent models, such as DeepSeek-R1, have also shown RLHF's potential to improve LLM reasoning on complex tasks. In RL, inference and training co-exist, creating dynamic resource demands throughout the workflow. Compared to traditional RL, RLHF further challenges training efficiency due to expanding model sizes and resource consumption. 
Several RLHF frameworks aim to balance flexible abstraction and efficient execution. However, they rely on serverful infrastructures, which struggle with fine-grained resource variability. As a result, during synchronous RLHF training, idle time between or within RL components often causes overhead and resource wastage.

To address these issues, we present \sys, the first scalable training framework for synchronous RLHF, built on serverless computing environments. \sys adapts to dynamic resource demands throughout the RLHF pipeline, pre-computes shared prefixes to avoid repeated computation, and uses a cost-aware actor scaling strategy that accounts for response length variation to find sweet spots with lower cost and higher speed. In addition, \sys assigns workloads efficiently to reduce intra-function imbalance and idle time. Experiments on both physical testbeds and a large-scale simulated cluster show that \sys achieves up to $1.35\times$ speedup and $44.8\%$ cost reduction compared to the state-of-the-art baseline.

\end{abstract}

\maketitle

\section{Introduction}

\noindent\textbf{Motivation.}
%
\LLMs have advanced rapidly and reshaped generative \AI.  
The \LLMs are first pre-trained on massive corpora with trillions of tokens from the internet to acquire broad knowledge~\cite{brown2020languagemodelsfewshotlearners}.  
They are then fine-tuned on high-quality, human-annotated instruction data to specialize for target tasks and domains.  
However, prior studies show that \LLMs can still behave improperly and may generate harmful or unwanted content~\cite{ouyang2022rlhf,zheng2023rlhfppo,dai2023saferlhfsafereinforcement}.  
To address these issues, recent work applies \RL optimization to \LLMs, aligning model behavior with human preferences through \RLHF.
\RLHF has been widely adopted in the latest {\LLM}s~\cite{openai2024gpt4technicalreport,ouyang2022rlhf,zheng2023rlhfppo,deepseekai2025deepseekr1incentivizingreasoningcapability,kimiteam2025kimik15scalingreinforcement} to enhance performance on challenging tasks such as mathematical problem solving~\cite{xiong2024iterativepreferencelearninghuman,ouyang2022rlhf}, answering scientific questions~\cite{hou2024does,Sheng2025hybridflow}, and code generation~\cite{dutta2024applyingrlaifcodegeneration,deepseekai2025deepseekr1incentivizingreasoningcapability}.
%
%
\RLHF training is significantly more resource-intensive and complex than traditional \RL. While both follow a similar staged training paradigm, existing \RL frameworks are not designed for the large-scale demands of \RLHF. 
As model sizes grow and training is distributed across multiple GPUs and machines, the overhead from data transmission and model weight synchronization becomes a major bottleneck, limiting overall efficiency.

\noindent\textbf{Limitations of state-of-art approaches.}
%
\RL separates data sampling from model updates, but the two processes remain interdependent. Model updates must wait for sampling to finish in order to use the most recent training data, while sampling is delayed by model updates because it needs synchronized model weights.
While traditional \RL training already suffers from significant resource idleness on serverful frameworks~\cite{liang2018rllibabstractionsdistributedreinforcement,mei2024srlscalingdistributedreinforcement}, \RLHF amplifies this inefficiency due to the demands of running \LLMs and heavy data transmission. 
Several works~\cite{slime_github,griggs2025skrylv01,li2026a3poacceleratingasynchronousllm,fu2025areallargescaleasynchronousreinforcement} attempt to improve utilization by converting synchronous \RLHF into asynchronous or off-policy execution. However, prior studies~\cite{zheng2025prosperitycollapsefaroffpolicy,noukhovitch2025asynchronousrlhffasterefficient,fu2025areallargescaleasynchronousreinforcement,li2026a3poacceleratingasynchronousllm} have shown that stale data introduced by asynchronous training can degrade convergence stability and final training quality. 
More recent synchronous \RLHF systems~\cite{Sheng2025hybridflow,hu2024openrlhfeasytousescalablehighperformance,zhong2024rlhfuseefficientrlhftraining}
improve efficiency through careful model placement and parallelism strategies, but they rely on serverful infrastructures with static resource allocation, which inherently suffer from unavoidable resource inefficiency and cannot dynamically adapt to the execution characteristics of \RLHF.
%
%
Serverless \RL frameworks~\cite{yu2024minionsrl,yu2024stellaris,yu2025nitro} shed light on this problem by demonstrating that dynamically launching and releasing components on demand can substantially reduce idleness and training costs. 
However, existing serverless solutions for traditional \RL cannot be directly applied to \RLHF systems,
%
and building a serverless \RLHF framework is challenging due to fundamental differences between \RLHF and traditional \RL, such as the model scale and transmission overhead mentioned above.
Moreover, the auto-regressive nature of \LLM generation and the continuously updated model weights lead to dynamic resource demands, both within a single step and across different training steps.

\noindent\textbf{Key insights and contributions.}
Existing \RLHF algorithms are highly sampling-heavy, where sequence generation dominates both execution time and resource consumption throughout training~\cite{shao2024grpo,yu2025dapoopensourcellmreinforcement,yue2025vapoefficientreliablereinforcement}.
By analyzing existing synchronous \RLHF workloads, we identify three key insights.
First, tight dependencies between sampling and learning, combined with repeated \KV cache computation during parallel generation, lead to substantial and unnecessary resource wastage.
Second, long-tail responses cause stragglers during sequence generation and leave allocated resources idle.
Third, workloads vary across training steps, as response lengths for the same prompt may increase or decrease over time, making static resource allocation inefficient.
Motivated by these observations, we present \sys, the first scalable serverless framework for synchronous \RLHF training, which improves efficiency and reduces cost.
Before each training step, \sys constructs an optimized generation plan.
It first analyzes the prompt dataset and training configuration to deduplicate \KV cache computation under parallel generation.
Next, \sys estimates the upcoming workload and dynamically scales the number of sampling actors\footnote{An actor is a separate component that hosts a model replica for response generation.} to balance training speed and resource cost.
\sys further groups prompts with similar predicted response lengths to mitigate stragglers and reduce intra-actor imbalance.
During execution, \sys places actors with locality awareness, prioritizing heavy workloads on nearby GPUs to better overlap communication and computation.
Our main contributions are summarized as follows:
\begin{itemize}[left=10pt,topsep=0pt,itemsep=0pt,parsep=0pt,partopsep=0pt]
    \item We present \sys, the first serverless training framework for \RLHF that reduces cost while accelerating training.
    \item We design a deduplicated prefill mechanism that eliminates redundant \KV cache computation in \RLHF.
    \item We propose a response-length prediction method with a cut-and-migrate backoff strategy to guide workload scheduling and actor scaling.
    \item We implement \sys on top of VERL~\cite{Sheng2025hybridflow} and evaluate it on both a physical cluster with two 8-GPU instances and a simulated large-scale cluster.
    Extensive experiments show that \sys reduces cost by up to 44.8\% and improves training speed by up to 1.35$\times$ compared to the state-of-the-art baseline.
\end{itemize}
\vspace{0.3\baselineskip}

\noindent\textbf{Experimental methodology and artifact availability.}
We implement \sys on top of VERL~\cite{Sheng2025hybridflow}, a widely used open-source \RLHF framework, and deploy it on a GPU cluster with two AWS EC2 \texttt{g6e.48xlarge} instances, each with 8 NVIDIA L40S GPUs, 192 AMD EPYC 7R13 vCPUs, and 384GB of system memory.
Additionally, we use Vidur~\cite{agrawal2024vidurlargescalesimulationframework} to run large-scale simulations.
We evaluate \sys using representative synchronous \RLHF algorithms, including \PPO~\cite{zheng2023rlhfppo} and \GRPO~\cite{shao2024grpo}, on standard benchmarks covering mathematics (GSM8K~\cite{cobbe2021gsm8k}), scientific reasoning (GPQA~\cite{rein2024gpqa}), and code generation (LiveCodeBench~\cite{jain2024livecodebench}).
Following prior work~\cite{romero2021llamaheterogeneousserverless,yu2024stellaris}, we report per-step execution time and GPU$\times$second cost to capture both performance and resource efficiency.
We will release the source code and experimental traces upon acceptance.

\noindent\textbf{Limitations of the proposed approach.}
\sys primarily focuses on synchronous \RLHF training.
%
While the actor scaling and prompt assignment designs in \sys are orthogonal and could be integrated into asynchronous \RLHF systems~\cite{fu2025areallargescaleasynchronousreinforcement,li2026a3poacceleratingasynchronousllm} with moderate adaptations, we leave such extensions to future work.

\section{Background and Motivation}
\label{sec:bg&motivation}

\subsection{\RLHF \vs General RL}

\begin{figure}[t]
  \centering
  \includegraphics[width=.9\linewidth]{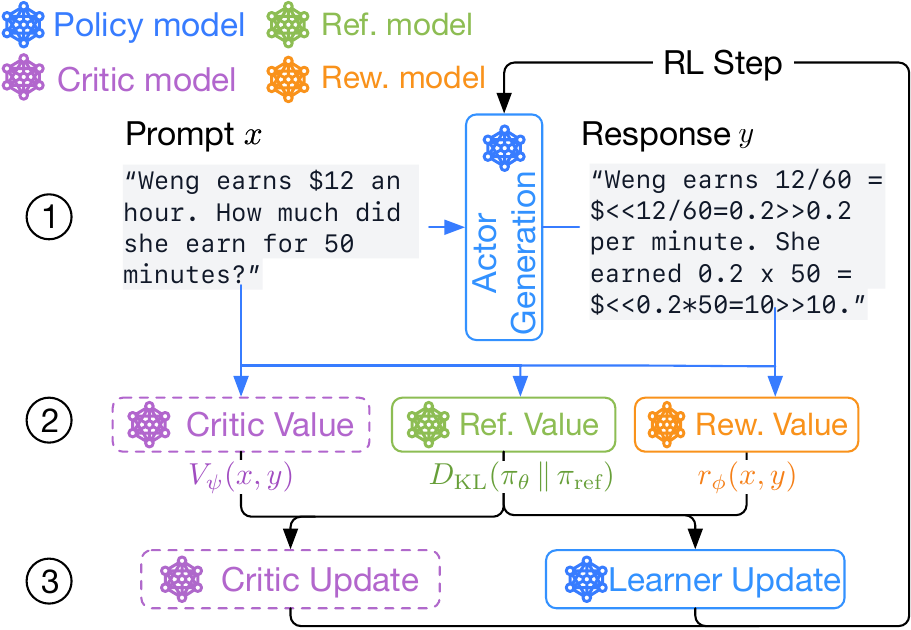}
  \vspace{-0.05in}
  \caption{\RLHF's dataflow in one training step, including the generation phase, preparation phase, and learning phase. Some details, like the use of the critic model, can vary depending on the specific algorithm used.
 } 
  \vspace{-0.1in}
  \label{fig:background}
\end{figure}

\Acf{RLHF} is a method used to align the behavior of a language model with human preferences by incorporating feedback into the training loop. 
Rather than relying solely on supervised data, \RLHF frames the alignment task as an \RL problem, where a policy model (i.e., the \LLM) $\pi_\theta$ is optimized to generate outputs that maximize the predicted reward provided by the pre-trained \RM $r_\phi(x, y)$.
\RLHF training contains multiple iterative training steps, and each training step involves three main phases, shown in Fig.~\ref{fig:background}:

\textbf{Phase-1 Generation:} For a batch of prompts $x$ from the prompt dataset, multiple sampling \textit{\textbf{actors}} will be launched using the policy $\pi_\theta$ to generate one or more candidate responses $y$. 
Each actor maintains its own replica of the policy model.

\textbf{Phase-2 Preparation:} The \RM $r_\phi(x, y)$ assigns a scalar score to each generated response.  
To prevent the policy from deviating too far from its original behavior, many algorithms~\cite{li2024remaxsimpleeffectiveefficient,zheng2023rlhfppo,shao2024grpo} incorporate a \KL divergence penalty between the current policy $\pi_\theta$ and the reference model $\pi_{\theta_0}$.

\textbf{Phase-3 Learning:} 
A \textit{\textbf{learner}}, equipped with a training engine~\cite{zhao2023pytorchfsdpexperiencesscaling,deepspeed}, is responsible for computing gradients and updating the policy model using the data produced from previous phases.

\textbf{Why existing \RL frameworks cannot fit?}
Existing \RL frameworks~\cite{liang2018rllibabstractionsdistributedreinforcement,yu2025nitro,yu2024minionsrl,yu2024stellaris,mei2024srlscalingdistributedreinforcement,Ustaran-Anderegg_AgileRL,MSRL} are not designed to support the scale and complexity of \RLHF for {\LLM}s. 
Traditional \RL workloads involve small models that can be trained on a small set of GPUs and rely on simple, synchronous architectures~\cite{liang2018rllibabstractionsdistributedreinforcement,MSRL,mei2024srlscalingdistributedreinforcement,yu2024minionsrl,yu2025nitro}. 
In contrast, \LLM training requires distributing models across multiple GPUs using techniques like data, tensor, and pipeline parallelism~\cite{Sheng2025hybridflow,hu2024openrlhfeasytousescalablehighperformance,zhao2023pytorchfsdpexperiencesscaling,deepspeed}, along with specialized inference~\cite{kwon2023vllm,shoeybi2020megatronlmtrainingmultibillionparameter,zheng2024sglangefficientexecutionstructured} and training~\cite{zhao2023pytorchfsdpexperiencesscaling,deepspeed} engines.
Moreover, scaling up the sampling actors in \RLHF increases the need for synchronized replicas, which leads to expensive and slow weight synchronization~\cite{Sheng2025hybridflow}. 
In such a case, various \RLHF-optimized frameworks~\cite{hu2024openrlhfeasytousescalablehighperformance,Sheng2025hybridflow,yao2023deepspeedchateasyfastaffordable,shen2024nemoalignerscalabletoolkitefficient,vonwerra2022trl} have been implemented to address those new challenges.

\subsection{Limitations of Serverful \RLHF Systems}
\label{subsec:limitation}

\RLHF faces several key challenges stemming from both its \RL nature and the use of {\LLM}s.  
The \RL loop inherits the resource inefficiency commonly observed in traditional \RL, caused by idle time between dependent components.  
For the \LLM, the policy model generates responses with different lengths, and is continually updated, leading to varying workloads and dynamic resource demands.
Existing \RLHF frameworks~\cite{hu2024openrlhfeasytousescalablehighperformance,Sheng2025hybridflow,slime_github} rely on serverful infrastructures and fail to address the issues both across the \RL loop and within individual \RLHF components.

\begin{figure}[t]
  \centering
  \includegraphics[width=.9\linewidth]{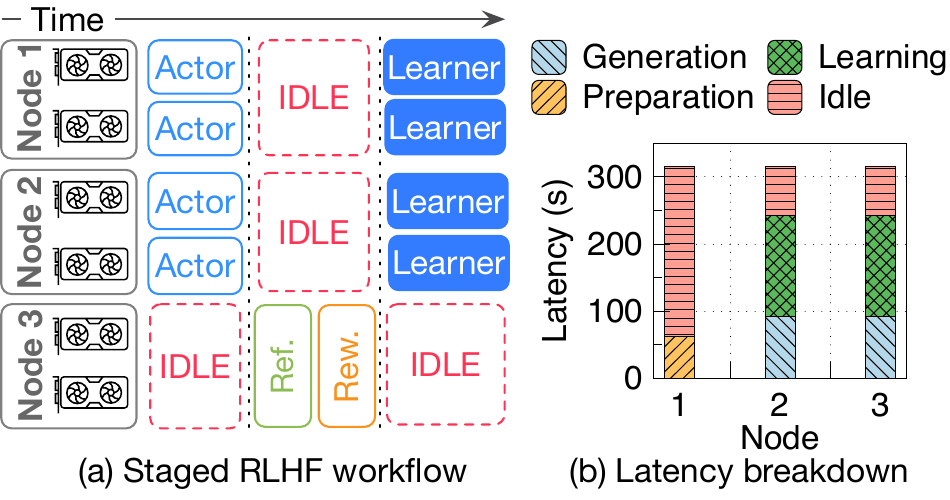}
  \vspace{-0.15in}
  \caption{(a) \RLHF's staged workflow causes idle time between components. (b) Idle components and (c) repeated calculation in \RLHF lead to resource wastage.} 
  \vspace{-0.1in}
  \label{fig:challenges_resource_wastage}
\end{figure}

\ricky{
\textbf{1) From the outer \RL loop: Idle components incur resource wastage on serverful platforms.} 
%
In this setup, components from different \RLHF phases have inter-dependencies, which often result in idle GPU resources during certain stages.
Fig.~\ref{fig:challenges_resource_wastage}(a) shows the workflow of \GRPO~\cite{shao2024grpo}, the default \RLHF algorithm in DeepSeek R1~\cite{deepseekai2025deepseekr1incentivizingreasoningcapability}, running in standalone placement where each model uses a separate set of GPUs. 
Fig.~\ref{fig:challenges_resource_wastage}(b) shows the latency breakdown in a training step for one of the \GRPO model placement solutions provided by VERL~\cite{Sheng2025hybridflow}, a commonly adopted \RLHF framework, using Qwen2.5-3B model~\cite{qwen2025qwen25technicalreport} with GSM8k dataset~\cite{cobbe2021gsm8k}.
While some \RLHF works~\cite{slime_github,griggs2025skrylv01,han2025asyncflowasynchronousstreamingrl,fu2025areallargescaleasynchronousreinforcement} attempt to address this issue by making the staged \RLHF work asynchronous, prior research has shown that asynchronous training can degrade final training quality due to the inconsistencies introduced by stale data~\cite{zheng2025prosperitycollapsefaroffpolicy,noukhovitch2025asynchronousrlhffasterefficient,fu2025areallargescaleasynchronousreinforcement,li2026a3poacceleratingasynchronousllm}.
Existing advanced synchronous \RLHF solutions~\cite{hu2024openrlhfeasytousescalablehighperformance,Sheng2025hybridflow,shen2024nemoalignerscalabletoolkitefficient} address this by optimizing model placement strategies---selectively co-locating different components on the same set of resources. However, this presents a difficult trade-off: placing components separately leads to GPU under-utilization, while co-locating them can easily cause \OOM errors, model context switching latency, or performance interference between components~\cite{Sheng2025hybridflow,shen2024nemoalignerscalabletoolkitefficient,zhong2024distservedisaggregatingprefilldecoding}.
}

\begin{figure}[t]
  \centering
  \includegraphics[width=0.9\linewidth]{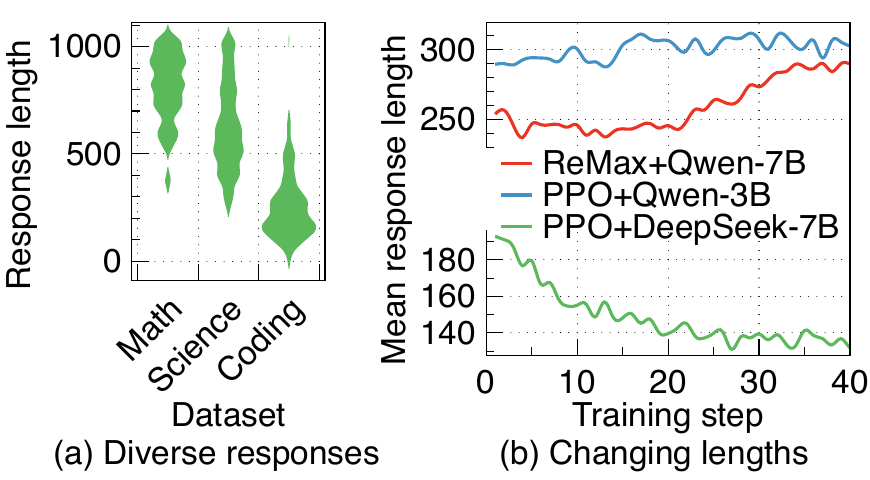}
  \vspace{-0.1in}
  \caption{Dynamic resource demands caused by: (a) significantly varying response lengths across and within different types of datasets, such as mathematical problems~\cite{aime2024}, general scientific questions~\cite{rein2024gpqa}, and coding tasks~\cite{jain2024livecodebench}; 
  (b) changing response lengths during \RLHF training.} 
  \vspace{-0.2in}
  \label{fig:challenges_dynamic_lengths}
\end{figure}

\textbf{2) Across training steps: Dynamic response lengths make static resource allocation inefficient.}
Recent studies have shown that during \RLHF's generation phase, the average response length for each training step keeps changing with an increasing or decreasing trend~\cite{deepseekai2025deepseekr1incentivizingreasoningcapability,singhal2024longwaygoinvestigating,kimiteam2025kimik15scalingreinforcement}. We run experiments with different \RLHF algorithms and model combinations using the GSM8k dataset to verify this pattern, shown in Fig.~\ref{fig:challenges_dynamic_lengths}(b). For most large-scale reasoning models, the response length will get longer as the training round increases~\cite{deepseekai2025deepseekr1incentivizingreasoningcapability,kimiteam2025kimik15scalingreinforcement}. Longer response lengths prolong inference time due to {\LLM}s' token-level auto-regressive generation, which indicates a dynamic resource demand for each training step.
Serverful platforms, with typically static resource allocation and long startup times, are unable to adapt to such dynamic resource demands.

\begin{figure}[t]
  \centering
  \includegraphics[width=0.9\linewidth]{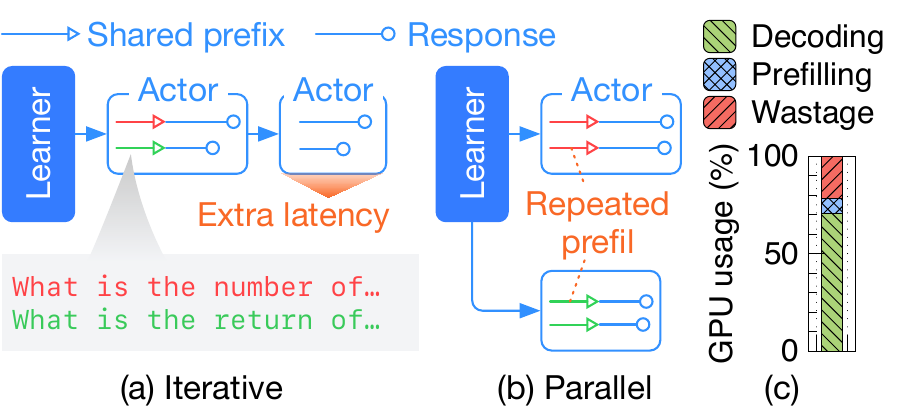}
  \vspace{-0.15in}
  \caption{\ricky{Existing generation strategies~\cite{Sheng2025hybridflow,hu2024openrlhfeasytousescalablehighperformance,zheng2024sglangefficientexecutionstructured,kwon2023vllm} for sampling-heavy \RLHF workloads, including (a) iterative generation, which introduces additional latency, and (b) parallel generation, which leads to (c) unnecessary \KV recalculation.}} 
  \vspace{-0.1in}
  \label{fig:motivation_deduplicated_prefill}
\end{figure}

\ricky{
\textbf{3) During each training step: Redundant \KV cache calculation causes unnecessary costs.}
When an \LLM generates a response, the prompt is first processed once to build the \KV cache during prefill, then tokens are produced using this cache during decoding~\cite{dai2019transformerxlattentivelanguagemodels,vaswani2023attentionneed}. Many \RLHF algorithms generate multiple responses per prompt~\cite{shao2024grpo,yu2025dapoopensourcellmreinforcement,yue2025vapoefficientreliablereinforcement,zheng2023rlhfppo}, and many datasets share a large portion of prefixes among prompts. As a result, the same or similar prefill computation is often repeated unnecessarily.
Prior \LLM serving works reduce this cost by reusing the \KV cache for prompts with shared prefixes~\cite{kwon2023vllm,liu2024cachegenkvcachecompression,xiao2024efficientstreaminglanguagemodels,sun2024cacheoncedecoderdecoderarchitectures}.
Some \RLHF systems, such as VERL~\cite{Sheng2025hybridflow}, apply this idea by splitting prompts into smaller batches, computing the \KV cache once, and reusing it across batches within the same training step, as shown in Fig.~\ref{fig:motivation_deduplicated_prefill}(a).
While this approach reduces duplicated prefill computation, it enforces sequential execution across batches.
Under static GPU resource allocation, this serialization increases the overall generation time and limits parallelism.
One may attempt to parallelize generation to improve throughput; however, this causes the same \KV cache to be recomputed independently by different parallel workers, leading to unnecessary GPU resource waste, as illustrated in Fig.~\ref{fig:motivation_deduplicated_prefill}(b).
Fig.~\ref{fig:motivation_deduplicated_prefill}(c) shows that approximately 22\% of the total GPU-time is spent on redundant \KV cache computation, when running \GRPO with Qwen2.5-3B on GSM8K and generating four responses per prompt.
}

\textbf{4) Inside the actors: Varying response lengths lead to GPU under-utilization.}
Sampling actors may consume significantly different amounts of compute resources due to the wide variation in response lengths. Within a single actor, some responses may terminate early after just a few decoding steps, while others may continue much longer, extending the generation phase even after most responses have completed~\cite{zhong2024rlhfuseefficientrlhftraining}. This imbalance leads to GPU under-utilization, as certain compute resources remain idle while waiting for the longest responses to finish. Fig.~\ref{fig:challenges_dynamic_lengths}(a) shows the imbalanced response lengths using the Qwen2.5-3B model, with 1024 maximal response length, in three different training datasets, including AIME2024~\cite{aime2024}, GPQA~\cite{rein2024gpqa}, and LiveCodeBench~\cite{jain2024livecodebench}.



\subsection{Opportunities For Serverless RLHF}

We propose the first serverless \RLHF system by identifying the following opportunities to address the issues mentioned in \S\ref{subsec:limitation}.

\textbf{Serverless can release idle components on time to mitigate wastage.}
Serverless elasticity and pay-as-you-go execution can lower training cost by launching components only when needed and promptly releasing them when idle, as shown in prior serverless \RL systems~\cite {yu2024minionsrl,yu2024stellaris,yu2025nitro}---motivating a serverless \RLHF design.

\textbf{Scaling actors to accommodate the changing resource demands.}
Serverless makes it feasible to adapt actor counts to shifting workloads and search for ``sweet spots'' where extra parallelism shortens step time enough to reduce total cost; achieving this requires a cost model to guide scaling decisions, and is harder on serverful platforms due to long startup latency~\cite{gcp_instance,aws_ec2,azure_vm}.

\textbf{Deduplicated prefill to mitigate repeated calculations in a serverless environment.}
By grouping prompts with shared prefixes ahead of time, and exploiting algorithms (\eg, \GRPO~\cite{shao2024grpo}) that sample multiple responses per prompt, we can compute prefill once and reuse it across responses---targeting redundant computation reduction (rather than serving throughput).

\textbf{Grouping prompts with similar lengths in the same actor to address workload imbalance.}
Because actor cost depends on execution time, a few long responses can keep an actor alive while other resources sit idle; predicting response lengths and co-locating similar-length prompts lets some actors finish sooner and be released earlier. 

\section{Objectives and Challenges}


We design a serverless \RLHF training system, \sys, to achieve three primary objectives:

\textbf{Speed up the end-to-end \RLHF training process.}  
Since the generation phase exhibits dynamic resource demands due to varying response lengths, we aim to accelerate the process by adjusting the number of actors to meet these changing resource needs.

\textbf{Mitigate redundant computation at each training step.}  
Shared prompt prefixes and multi-response generation in many \RLHF algorithms~\cite{yu2025dapoopensourcellmreinforcement,shao2024grpo,yue2025vapoefficientreliablereinforcement} cause repeated KV cache computation during generation, leading to unnecessary cost. We aim to reduce this overhead using a deduplicated prefill.

\textbf{Improve resource efficiency inside \RL components.}  
The response lengths assigned to a single actor can vary significantly, and outlier-long responses can prolong the actor's execution time while leading to low GPU utilization. Our goal is to reduce resource wastage caused by imbalanced workloads within each actor.

To achieve the goals, three challenges must be addressed:

\textbf{How to trade off between speedup and cost?}  
Scaling up the number of actors can speed up the generation phase, but usually increases resource cost. To find a balance, or even a sweet spot, with minimal cost and shortest time, it's necessary to estimate both cost and execution time accurately.

\textbf{How to improve prefill efficiency in \RLHF?}  
Using a deduplicated prefill actor with fewer resources can reduce cost by computing each prompt only once. However, allocating too few resources may slow down the prefill stage, introducing latency into the overall \RLHF training process.

\textbf{How to minimize idle time within each actor?}  
Grouping prompts with similar response lengths into the same decode actor can reduce GPU under-utilization and save cost. To do this effectively, we need an estimate of response lengths prior to assignment.

\section{\protect\sys's Design}

\subsection{Overview}

\begin{figure}[t]
  \centering
  \includegraphics[width=0.9\linewidth]{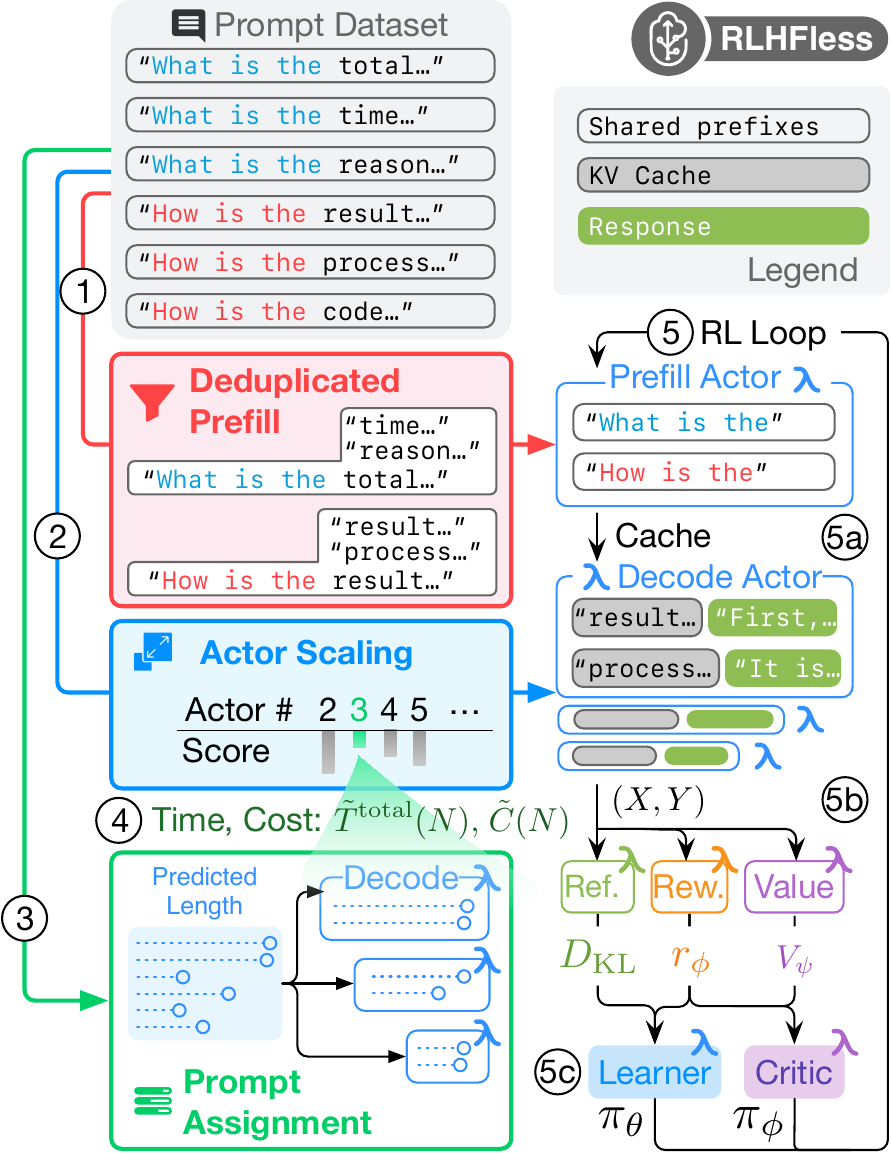}
  \vspace{-0.05in}
  \caption{\sys's overview. } 
  \vspace{-0.15in}
  \label{fig:design_overview}
\end{figure}

\sys is the first \RLHF training system designed for serverless platforms. It leverages the elasticity and fine-grained resource management of serverless computing to facilitate \RLHF training. 
Fig.~\ref{fig:design_overview} illustrates \sys's architecture and workflow.
%
\sys executes five main steps in every training step:

\textbf{Step~{\Large \circled{\small 1}}: Setting up the prefill actor.}  
\sys analyzes the prompt dataset or algorithm-specific hyperparameters to identify a deduplicated batch of shared prefixes. These shared prompts are assigned to a dedicated prefill actor, which computes the initial KV cache to be reused by all decode actors during the decoding stage.

\textbf{Step~{\Large \circled{\small 2}}:
Iterating all available actor numbers.}  
\sys employs a cost-aware planning module to estimate both execution time and total cost across a range of possible actor counts, with the corresponding prompt assignment strategies. 

\textbf{Step~{\Large \circled{\small 3}}:
Prompt assignment among decode actors.}  
When calculating the scaling score with a given actor number and workload, \sys first generates a corresponding prompt assignment plan for better cost and time estimation.
\sys utilizes historical training data to predict the response length of each prompt in the current training batch. It then sorts prompts by estimated length and evenly distributes them into the $N'$ decode actors, grouping prompts with similar lengths to minimize resource wastage and improve actor efficiency.

\textbf{Step~{\Large \circled{\small 4}}:
Scaling the decode actors.}  
The prompt assignment module returns the workload assignment plan given the actor number and its corresponding execution time and cost estimation.
Based on those results, \sys's actor scaling module selects the optimal number of actors $N'$ and its corresponding workload assignment plan by balancing speedup with resource efficiency.


\textbf{Step~{\Large \circled{\small 5}}: Executing \RLHF loop.}
After generating the execution plan for the generation phase of the next training step, \sys initiates the step and executes the three \RL phases, including {\Large \circled{\small 5a}} generation, {\Large \circled{\small 5b}} preparation, and {\Large \circled{\small 5c}} learning, following the structure of conventional \RLHF systems. 
The training process repeats the five steps until completion.

\subsection{Deduplicated Prefill}
\label{subsec:disaggregated_prefill}

\begin{figure}[t]
  \centering
  \includegraphics[width=0.8\linewidth]{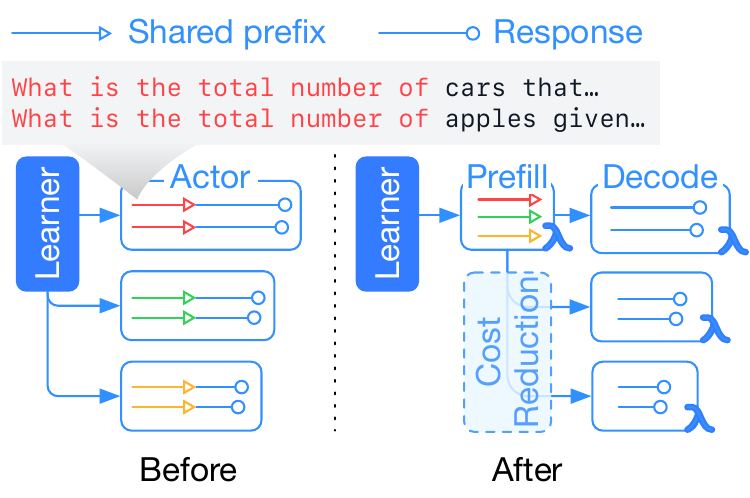}
  \vspace{-0.1in}
  \caption{\sys's deduplicated prefill design.} 
  \vspace{-0.1in}
  \label{fig:design_disaggregate_prefill}
\end{figure}


\RLHF algorithms~\cite{shao2024grpo,yu2025dapoopensourcellmreinforcement,yue2025vapoefficientreliablereinforcement,ahmadian2024basicsrevisitingreinforcestyle,hu2025reinforce++} have duplicated KV cache computations in the generation phase of a training step due to two reasons:
1) shared prefixes among prompts in the training dataset, and  
2) multiple responses generated for a single prompt.  
As a result, the number of unique KV cache computations required in the prefill stage is often less than the number of response generations performed in the decode stage.  
Existing serverful \RLHF frameworks~\cite{hu2024openrlhfeasytousescalablehighperformance,Sheng2025hybridflow} ignore this characteristic and allocate static resource sizes throughout training. 
In contrast, serverless computing offers a pay-as-you-go model, allowing us to reduce cost by applying a deduplicated prefill-decode design. This enables allocating only the necessary resources for the prefill actor and computing each unique KV cache exactly once.  
Since the prefill actor is compute-bound, similar to the learner~\cite{Sheng2025hybridflow,zhong2024distservedisaggregatingprefilldecoding}, we unify resource allocation and model parallelism strategy for both components.

We next estimate its maximum prefill batch size $B_{\text{prefill}}$ using a profiling-based capacity model.
\sys then chooses a target shared-prefix length $L$ so that the prefill actor is filled to capacity with deduplicated prompts.
Concretely, let $X$ be the training prompt set and define
a candidate prefix length $L\in[L_{\min},L_{\max}]$, where
\begin{align*}
    L_{\min}=\min_{x\in X}\|x\|,\quad L_{\max}=\max_{x\in X}\|x\|.
\end{align*}
For a candidate prefix length $L$, let $D(L)$ be the number of unique prefixes of length $L$ in $X$ (\ie, the count of deduplicated prompts at prefix length $L$).
\sys sweeps $L\in[L_{\min},L_{\max}]$ and selects the largest $L$ such that
\[
D(L)\le B_{\text{prefill}}.
\]
As $L$ increases, $D(L)$ decreases; when $D(L)$ first satisfies $D(L)\le B_{\text{prefill}}$, the prefill actor is fully utilized and computes the maximum amount of deduplicated KV cache within its capacity.

\subsection{Prompt Assignment}
\label{subsec:prompt_assignment}

\begin{figure}[t]
  \centering
  \includegraphics[width=0.98\linewidth]{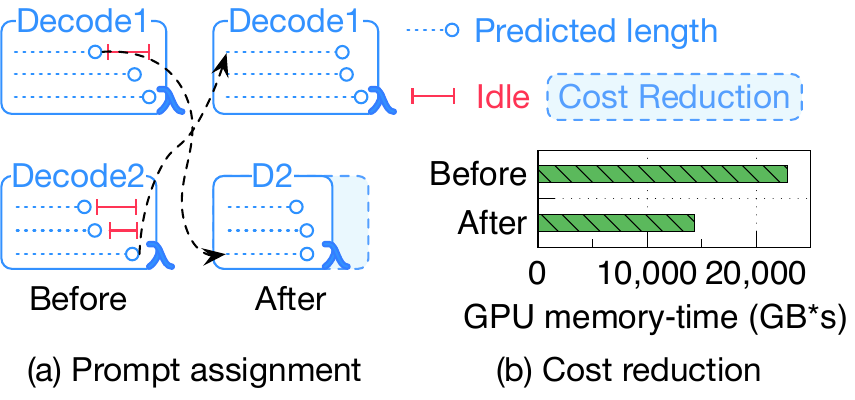}
  \vspace{-0.1in}
  \caption{(a) \sys's prompt assignment design, and (b) the potential benefits that can be achieved under \GRPO training with Qwen2.5-3B~\cite{qwen2025qwen25technicalreport} model and GSM8k~\cite{cobbe2021gsm8k} dataset.
  } 
  \vspace{-0.1in}
  \label{fig:design_prompt_assignment}
\end{figure}

Due to the varying lengths of responses generated within a single sampling actor, GPU utilization decreases when most sequences finish early and only a few long responses remain.
Previous work~\cite{zhong2024rlhfuseefficientrlhftraining} proposes a cut-and-migrate strategy to interrupt sequence generation in the middle, and migrate all unfinished responses to one actor to save resources for other tasks (\eg, tasks in preparation phase).
However, such a cut-and-migrate method introduces extra overhead due to the data transmission when migrating unfinished responses from one actor instance to another.
Moreover, this method interrupts all actors simultaneously, harvesting resources only in a coarse-grained manner and conflicting with the fine-grained resource management principles of serverless computing.

To enable efficient cut-and-migrate in serverless environments, we propose a novel dynamic ranking-based prompt assignment strategy shown in Fig.~\ref{fig:design_prompt_assignment}(a), which significantly reduces the transmission cost by minimizing the number of cut responses.
This approach releases idle resources at a finer granularity by allowing actors to terminate at different times, while keeping the workload balanced within each actor.
As shown in Fig.~\ref{fig:design_prompt_assignment}(b), grouping prompts with similar predicted response lengths into the same actor can reduce this inefficiency by allowing early termination for each actor, thereby saving GPU time and cost.
To enable this, we require a response length predictor to estimate workloads ahead of time. 




\textbf{Response length estimation.}
Accurately estimating response generation workloads in advance is essential for assigning prompts to actors in a cost-efficient way. Specifically, we need to predict the expected response length for each prompt before launching the actors that will process it.
Several recent works have proposed response length prediction methods to improve \LLM serving efficiency~\cite{fu2024efficientllmschedulinglearning,qiu2024efficientinteractivellmserving,zheng2023responselengthperceptionsequence}. However, these approaches typically require either finetuning the \LLM to perform self-prediction~\cite{zheng2023responselengthperceptionsequence} or training an external proxy model~\cite{fu2024efficientllmschedulinglearning,qiu2024efficientinteractivellmserving}. Both options incur non-trivial training and prediction costs, adding complexity by introducing an extra training phase for the predictor.
In addition, response length prediction remains inherently difficult due to the stochastic nature of \LLM outputs. Even \SOTA methods that improve throughput for \LLM serving still suffer from limited token length prediction accuracy~\cite{fu2024efficientllmschedulinglearning,qiu2024efficientinteractivellmserving,zheng2023responselengthperceptionsequence}.

A key advantage in the context of \RLHF is that we have access to both the training dataset and historical response length data from earlier training steps. Unlike \LLM serving, where predictions must be made from scratch for each user input, \RLHF training involves repeated sampling over a fixed set of prompts.
Moreover, our system design does not require exact response length estimates; rather, it only depends on the relative ranking of prompts by expected response length. As long as this ranking remains relatively stable across steps, we can make effective assignment decisions.
To validate this, we track the response length changes for each prompt throughout the \PPO training process on the GSM8K dataset, using the Qwen2.5-3B model with a maximum token limit of 2048.
%
Most response lengths exhibit consistent trends over time and do not fluctuate drastically between steps. 
Approximately 70\% of the response lengths vary by no more than 50 tokens across two consecutive epochs, and around 90\% change by no more than 100 tokens per epoch.
This suggests that we can use the historical response lengths observed in the previous training steps as a reliable estimator for the current step, eliminating the need for costly model-based predictors~\cite{zheng2023responselengthperceptionsequence,fu2024efficientllmschedulinglearning,qiu2024efficientinteractivellmserving}.
As for the first epoch without historical length data, we can make use of the ground truth answers from the training dataset as a temporary estimation.
We adopt the common \EWMA algorithm to take historical data as input and output the length estimation for each prompt. 
%


\begin{figure}[t]
  \centering
  \includegraphics[width=0.75\linewidth]{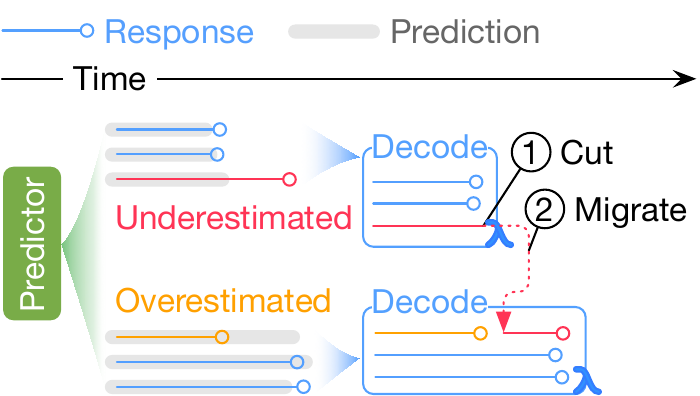}
  \caption{\sys's cut-and-migrate strategy. \sys cuts unfinished underestimated responses and migrates them to available decode actors with open slots, which have been released by completed, overestimated responses.} 
  \vspace{-0.2in}
  \label{fig:design_backoff}
\end{figure}

\textbf{Cut-and-migrate in \sys.}
Even with reasonably accurate response length estimation, mis-assignments can still occur. For example, if prompts with long actual responses are incorrectly assigned to actors intended for short responses, the resulting execution delay can reduce overall cost efficiency.  
To address this issue, we propose a novel cut-and-migrate strategy. Unlike existing methods that terminate all actors' executions simultaneously, \sys dynamically corrects prediction errors during execution, preserving the benefits of prompt assignment.

We classify the prediction errors into two categories: 1) Over-estimation: When a prompt with a short actual response is placed in an actor designed for long responses, GPU is under-utilized as the prompt finishes early and the remaining time is wasted.  
2) Under-estimation: When a prompt with a long response is placed in an actor assigned to short responses, it prolongs the actor execution time.
To mitigate the impact of prediction errors, we apply a cut-and-migrate strategy, inspired by existing work~\cite{zhong2024rlhfuseefficientrlhftraining}, shown in Fig.~\ref{fig:design_backoff}. During execution, if an actor is delayed due to long-running prompts and there are other available running actors, we terminate those generations and migrate them to the actors with available slots, which are released when over-estimated responses finish generation. This approach helps reduce tail latency and improves overall GPU utilization.
The timing to terminate an actor is controlled by a cutting threshold $\tau \in (0, 1]$. An actor is terminated when $\tau \times 100\%$ of its assigned responses have completed generation. This threshold balances the trade-off between waiting for stragglers and minimizing resource wastage, allowing the system to release most of the GPU resources while offloading the remaining long responses to other actors.

\subsection{Cost-Aware Actor Scaling}
\label{subsec:actor_scaling}

The average response length varies throughout \RLHF training, causing dynamic execution times across training steps in traditional \RLHF frameworks.  
With static resource allocation, this either results in resource wastage when fewer resources are needed or slow decoding when long responses cannot be processed efficiently.  
Thus, even if existing serverful \RLHF systems can optimize generation strategy, they cannot consistently reach a sweet spot of low cost and fast speed, since workloads change at each step while resources and actor numbers remain fixed.  
This highlights the opportunity to dynamically adjust the number of actors to match changing resource demands.

Previous studies~\cite{yu2025nitro,yu2024minionsrl} have demonstrated the potential of serverless computing's elasticity for actor scaling in traditional \RL training, highlighting the possibility of implementing cost-aware actor scaling in a serverless \RLHF system.  
However, scaling the number of actors in \RLHF differs from traditional \RL. Previous actor scaling methods either rely on a pretrained scheduler~\cite{yu2024minionsrl}, which introduces additional costs, or utilize the Hessian matrix of the policy model to assess data needs~\cite{yu2025nitro}. The latter approach can incur significant computation overhead as the policy model size increases to billions of parameters.  
Furthermore, in typical \RL, actor scaling alters the volume of training data by adjusting the number of actors, while in \RLHF, changing the number of actors primarily adjusts the allocated GPU resources during the generation phase to suit the changing resource demands at each training step.

It has been well established in prior work that the generation phase is memory-bound, which can achieve higher performance with increased data parallelism by deploying more model replicas to process prompts concurrently~\cite{zhong2024distservedisaggregatingprefilldecoding,li2023alpaservestatisticalmultiplexingmodel,Sheng2025hybridflow,zhong2024rlhfuseefficientrlhftraining}.  
In \RLHF, each sampling actor maintains a copy of the policy model, meaning that the number of actors effectively determines the data parallelism size. As a result, increasing data parallelism directly corresponds to scaling up the number of actors.
However, the speedup achieved by scaling up the number of actors may come at the cost of significantly increased resource consumption, which will further lead to exponential growth in training costs. This indicates the need for a cost-aware actor scaling design.

\textbf{Balanced speedup with lower final costs.} 
Estimating resource consumption before scaling up the number of actors is essential to avoid introducing unnecessary costs.  
The total cost for the $N$ decode actors, denoted as $C_{\text{Decode}}$, can be estimated as
\begin{equation*}
    C_{\text{Decode}} := \rho \sum_{i=1}^N T(X_i, G_i) \times G_i,
\end{equation*}
where $\rho$ is the unit cost of GPU usage, measured in dollars per GPU per unit time. $X_i$ represents the set of prompts assigned to the $i$th decode actor, and $G_i$ denotes the number of GPUs allocated to it. $T(X_i, G_i)$ is the estimated execution time for processing workload $X_i$ using $G_i$ GPUs.
To estimate the execution time $T(X_i, G_i)$, we leverage profiled data collected using dummy batches prior to training. Specifically, we measure the \TPOT latencies in the decode step, under different maximum response lengths, following the approach used in prior work~\cite{zhong2024distservedisaggregatingprefilldecoding,chittyvenkata2024llminferencebenchinferencebenchmarkinglarge}.  
For each prompt $x \in X_i$, the corresponding sequence length is estimated using the response length prediction module described in \S\ref{subsec:prompt_assignment}.

The end-to-end execution time for $N$ decode actors, denoted as $T^{\text{total}}(N)$, is determined by the slowest (i.e., longest-running) actor and is computed as
\begin{equation*}
    T^{\text{total}}(N) := \max_{i \in [1, N]} T(X_i, G_i).
\end{equation*}

For all candidate actor counts $N \in [N_{\min}, N_{\max}]$, where $N_{\min}$ and $N_{\max}$ represent the predefined minimum and maximum number of decode actors, respectively, we normalize the execution time and cost to obtain
\begin{equation*}
    \tilde{T}^{\text{total}}(N) := \frac{T^{\text{total}}(N) - T^{\text{total}}_{\min}}{T^{\text{total}}_{\max} - T^{\text{total}}_{\min}}, \quad
    \tilde{C}(N) := \frac{C_{\text{Decode}}(N) - C_{\min}}{C_{\max} - C_{\min}},
\end{equation*}
where $T^{\text{total}}_{\min}$ and $T^{\text{total}}_{\max}$ are the minimum and maximum execution times observed over the range of $N$, and $C_{\min}$ and $C_{\max}$ are the corresponding cost bounds.
To determine the optimal number of actors $N'$, we formulate the objective as a weighted sum of the normalized time and cost:
\begin{equation*}
    N' := \mathop{\arg\min}\limits_{N \in [N_{\min}, N_{\max}]} \left\{ 
    \lambda \cdot \tilde{T}^{\text{total}}(N) + (1 - \lambda) \cdot \tilde{C}(N)
    \right\},
\end{equation*}
where $\lambda \in [0,1]$ is a user-defined weight controlling the trade-off between execution time and cost. A larger $\lambda$ prioritizes faster execution, while a smaller $\lambda$ emphasizes lower cost.

\subsection{Locality-Aware Actor Placement}

\begin{figure}[t]
  \centering
  \includegraphics[width=0.75\linewidth]{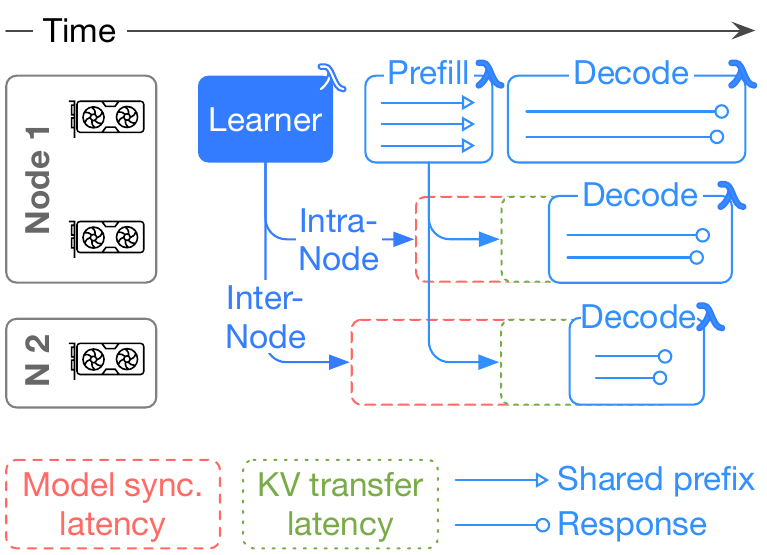}
  \vspace{-0.1in}
  \caption{\sys's locality-aware actor placement.} 
  \vspace{-0.2in}
  \label{fig:design_locality_aware}
\end{figure}

The model synchronization overhead is one of the primary bottlenecks in \RLHF training~\cite{zhong2024rlhfuseefficientrlhftraining,Sheng2025hybridflow}. This overhead occurs between the learning phase of the previous step and the generation phase of the current step.  
During synchronization, model weights are transmitted from the learner to all sampling actors, introducing unavoidable transmission latency.  
Furthermore, applying \sys's actor scaling strategy to increase the number of actors, along with the disaggregated prefill design that separates actors into a prefill actor and multiple decode actors, can further amplify data transmission overhead due to increased communication.  
To address both model synchronization latency and KV transfer delays, \sys incorporates a locality-aware actor placement strategy, which combines the prompt assignment mechanism to properly place the actors, minimizing the end-to-end execution time in each training step.

As discussed in \S\ref{subsec:prompt_assignment}, \sys leverages prompt assignment to group prompts with similar expected workloads into the same decode actor.  
In this setting, the execution time of each actor is determined by the prompt with the longest response length within that group. Consequently, the end-to-end generation time for a training step is dictated by the actor with the longest generation workload.  
To optimize for this, we first co-locate the prefill actor with the learner on the same set of GPU resources. This reduces the transmission overhead for model weight synchronization between the learner and the prefill actor. 
Additionally, since both the learner and the prefill actor are compute-bound components that benefit from a larger model parallelism size~\cite{Sheng2025hybridflow,zhong2024distservedisaggregatingprefilldecoding}, co-locating them does not require changes to the existing parallelism strategy.
Next, we co-locate the decode actor handling the longest response with the prefill actor, allowing it to start immediately without incurring latency for model weight and KV cache transfer.  
For additional decode actors introduced by actor scaling (\S\ref{subsec:actor_scaling}), we prioritize those with heavier workloads for placement on GPU nodes that are physically closer to the learner and prefill actor (\eg, within the same GPU node). Lighter workloads are assigned to more distant GPU resources.  
Since model weight synchronization can begin immediately after the learner finishes updating the model, while KV cache transfer can only start after the prefill stage completes, we can overlap transmission with execution. As illustrated in Fig.~\ref{fig:design_locality_aware}, transmission latency can be effectively hidden as long as the following condition holds for all $i \in \{2, \dots, N\}$:
\begin{equation*}
    L_{\text{Model}}^i + L_{\text{KV}}^i + L_{\text{Decode}}^i \leq L_{\text{Prefill}} + L_{\text{Decode}}^1,
\end{equation*}
where $N$ is the total number of decode actors, $L_{\text{Model}}^i$ and $L_{\text{KV}}^i$ denote the model synchronization and KV cache transfer latency for the $i$th decode actor, respectively.  
$L_{\text{Decode}}^i$ is the execution time of the $i$th decode actor, and $L_{\text{Decode}}^1$ is the execution time of the decode actor co-located with the learner and prefill actor.  
$L_{\text{Prefill}}$ represents the execution time of the prefill actor.
We add this constraint when scaling the actors to achieve a better actual performance, mitigating prolonged execution due to a larger-scale data transfer.

\section{Implementation}

We build \sys on top of an early version of VERL~\cite{Sheng2025hybridflow}, one of the most commonly adopted \RLHF frameworks, with integration of vLLM~\cite{kwon2023vllm} as the inference engine and \FSDP~\cite{zhao2023pytorchfsdpexperiencesscaling} as the training engine.

\noindent\textbf{Serverless cluster.}  
We build a serverless cluster using Docker~\cite{merkel2014docker}, a common approach in existing serverless environments. The Ray framework~\cite{moritz2018raydistributedframeworkemerging} manages compute resources for the serverless functions and coordinates communication among them.

\noindent\textbf{Serverless functions.}  
To mitigate the cold-start latency in the serverless \RLHF system, we adopt an invocation prediction method, similar to prior work~\cite{shahrad2020serverless,poppe2022moneyball}. Functions are prewarmed based on historical calls, using the shortest execution time from earlier steps plus a fixed buffer, ensuring they are ready before actual execution.

\noindent\textbf{Model configuration.}  
By default, \sys allocates the same number of GPUs to all decode actors.  
For internal settings (\eg, tensor or pipeline parallelism), we follow VERL’s documentation and tune hyperparameters to fit our GPUs.

\noindent\textbf{Workload estimation.}  
\sys’s core features require estimates of inference throughput and data transmission cost under specific parallelism and model sizes. We follow methods from prior work~\cite{agrawal2024vidurlargescalesimulationframework,zhong2024distservedisaggregatingprefilldecoding,thor2025vram} and incorporate profiling results from dummy batches run during training initialization.

\noindent\textbf{Data transmission.}  
Redis~\cite{redis} is used as an external distributed cache for data like historical response lengths, model checkpoints, and training traces. For model synchronization and \KV cache transfer, we use \NCCL~\cite{nccl} to avoid serialization overhead and enable direct communication.

\noindent\textbf{Locality-aware function placement.}  
To enable locality-aware placement, we profile GPU bandwidth and topology on the hardware testbeds. \sys uses these profiles with workload estimates to calculate cache transfer latencies. Model synchronization latency is relatively stable and can be estimated from historical results.

\section{Evaluation}


\subsection{Experimental Setup}

\noindent\textbf{Workloads and datasets.}
To comprehensively demonstrate the performance benefits of \sys, we integrate multiple algorithms and datasets, and conduct experiments across various combinations of algorithms, datasets, and model sizes.
Our algorithm choices include the widely adopted \PPO~\cite{zheng2023rlhfppo}, commonly used in early-stage \RLHF, as well as \GRPO~\cite{shao2024grpo}, which replaces the critic model with repeated generations and forms the basis of many state-of-the-art \RLHF algorithms~\cite{yue2025vapoefficientreliablereinforcement,yu2025dapoopensourcellmreinforcement}.
The training datasets span multiple domains, including the GSM8k mathematical reasoning dataset~\cite{cobbe2021gsm8k}, the GPQA scientific question dataset~\cite{rein2024gpqa}, and the LiveCodeBench coding benchmark~\cite{jain2024livecodebench}.
All models running on physical testbeds are based on the Qwen2.5 architecture~\cite{qwen2025qwen25technicalreport}, with tested sizes ranging from 3B to 7B.
Due to budget constraints, we run large-scale simulations for larger Llama models~\cite{grattafiori2024llama3herdmodels,touvron2023llama2openfoundation} using Vidur~\cite{agrawal2024vidurlargescalesimulationframework}.
The maximal prompt length and response length are set to 1024 and 2048, respectively.
We run each setting for three epochs and report the average per-step performance.

\noindent\textbf{Hardware and system environment.}
We use two AWS EC2 \texttt{g6e.48xlarge} instances as our hardware testbed. Each instance has 8 NVIDIA L40S Tensor Core GPUs with 384GB VRAM in total, 192 AMD EPYC 7R13 vCPUs, and 384 GB CPU memory.  

\noindent\textbf{Software stack and configuration.}
The operating system of the testbeds is Ubuntu 22.04. The NCCL and CUDA versions are 2.27 and 12.8, respectively. We build \sys based on VERL 0.3.1.dev, and the selected inference engine is vLLM 0.8.3.

\noindent\textbf{Baselines and comparisons.}
VERL~\cite{Sheng2025hybridflow} is a widely used open-source \RLHF framework, and we build \sys on top of it.
The core designs of \sys
are general and can be integrated into other synchronous \RLHF systems~\cite{hu2024openrlhfeasytousescalablehighperformance,slime_github,shen2024nemoalignerscalabletoolkitefficient}.
To ensure a fair comparison, we use VERL as our primary baseline.
We additionally include RLHFuse~\cite{zhong2024rlhfuseefficientrlhftraining} as a baseline to evaluate the effectiveness of prompt assignment in \S\ref{subsec:effectiveness}.
For the length prediction comparison in \S\ref{subsec:sensitivity}, we train a model-based predictor following prior work~\cite{qiu2024efficientinteractivellmserving,fu2024efficientllmschedulinglearning}.

\noindent\textbf{Methodology and experimental protocol.}
We record the execution time (in seconds) and the cost of each training step, and report the average per-step values over the entire training process.
For each serverless function invocation, resource consumption is computed as the product of allocated resources and execution time.
In practice, we measure GPU$\times$second and use it as the cost proxy for each serverless function, following prior work~\cite{yu2024stellaris,romero2021llamaheterogeneousserverless}.

%

\noindent\textbf{Reproducibility kit (AD/AE).}
The source code of \sys and the experimental traces will be publicly released if accepted.

%

\noindent\textbf{\sys's setting.}  
Considering our resource limits and the model sizes used in experiments, each actor is allocated two GPUs for 3B training and four GPUs for 7B training.
For the serverful baseline, we set the actor number to 3, as profiling shows this provides the best speed-cost balance under static resource allocation. More active actors for the entire training would significantly increase the cost.
%
%
For the \RLHF training hyperparameters in both \sys and the baseline, we follow the configurations provided by VERL~\cite{Sheng2025hybridflow}, while tuning key parameters such as batch size and KV cache allocation rate to match our hardware testbed across different experimental setups.  
For the hyperparameters of \sys's core components, we set the \EWMA window size to 1, the cut-and-migrate threshold $\tau$ to 0.7, and the actor scaling weight $\lambda$ to 0.7. These choices are justified and further analyzed in \S\ref{subsec:sensitivity}.

\subsection{Overall Performance}
\label{subsec:overall_performance}


\begin{figure}[t]
  \centering
  \includegraphics[width=0.85\linewidth]{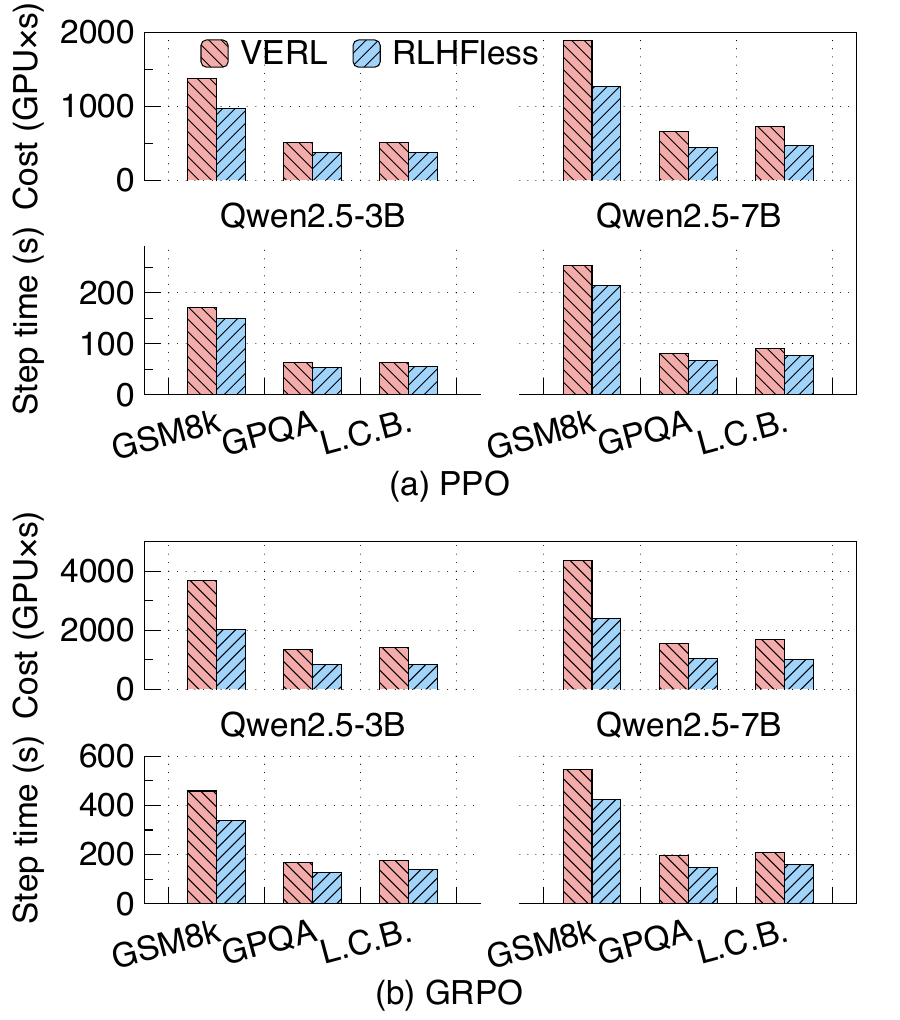}
  \vspace{-0.15in}
  \caption{\sys's overall performance evaluation.} 
  \label{fig:eval_overall}
\end{figure}

We evaluate how \sys improves \RLHF generation efficiency in terms of both speed and cost, by comparing it with a naive version of \sys built on VERL~\cite{Sheng2025hybridflow}, where all core features are disabled.  
Fig.~\ref{fig:eval_overall} shows that \sys speeds up \RLHF training and lowers cost by (1) dynamically scaling actors to match changing resource demands, (2) eliminating repeated computation, and (3) releasing idle resources promptly.  
Training on GSM8k takes the longest, due to many long responses, which is consistent with the profiled results in Fig.~\ref{fig:challenges_dynamic_lengths}(a).  
\sys delivers larger gains on \GRPO than on \PPO because \GRPO is more sampling-heavy, leaving more room to reduce waste during generation.  
We also observe that as training scale and workload grow, the benefits of \sys increase, suggesting its potential to adapt and deliver greater gains in super-large \RLHF workloads.
Overall, \sys speeds up training by up to $1.35\times$ and reduces cost by up to $44.8\%$.


\subsection{Ablation Study}
\label{subsec:ablation}

\begin{figure}[t]
  \centering
  \includegraphics[width=0.9\linewidth]{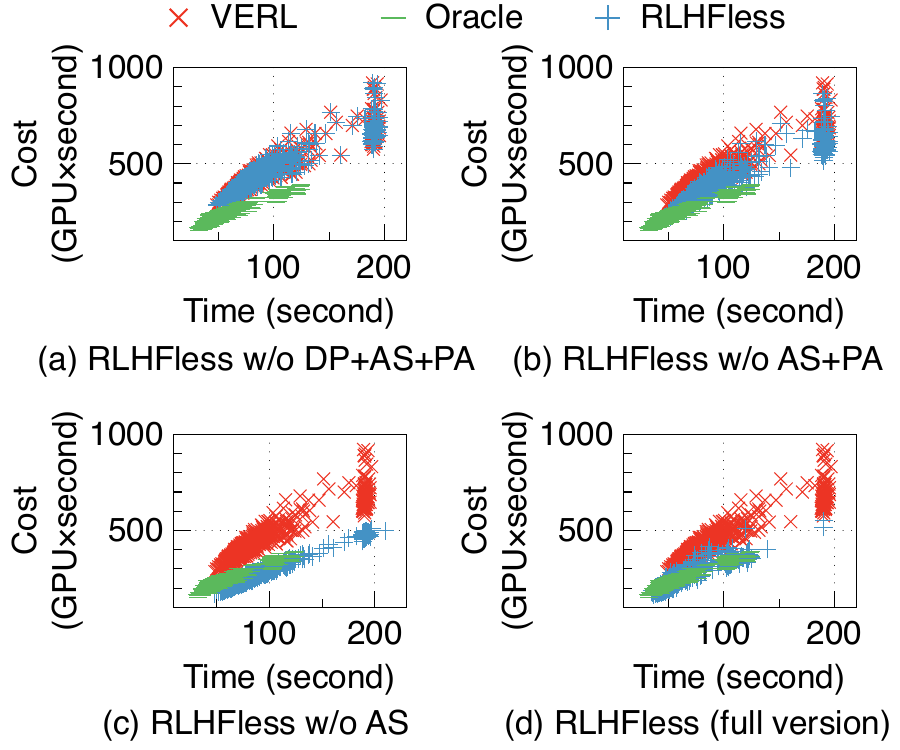}
  \vspace{-0.1in}
  \caption{\sys's ablation study (DP: \underline{D}eduplicated \underline{P}refill; AS: \underline{A}ctor \underline{S}caling; PA: \underline{P}rompt \underline{A}ssignment). Each point shows the GPU cost and execution time of one training step.
Points closer to the lower-left corner indicate lower cost and faster execution.
Subfigure (a) shows \sys with all core features disabled.
From (b) to (d), we enable one additional core feature at a time, illustrating the incremental benefits of each design.}
  \vspace{-0.2in}
  \label{fig:eval_ablation}
\end{figure}

We evaluate the effectiveness of \sys's core features---deduplicated prefill, prompt assignment, and actor scaling---via an ablation study using several system variants. We run experiments using the \GRPO algorithm and the Qwen2.5-3B model on the GSM8k dataset.
The study considers four variants of \sys: 1) \sys \textit{w/o DP+AS+PA}: raw \sys based on VERL, with no deduplicated prefill, no prompt assignment, and no actor scaling; 2) \sys \textit{w/o AS+PA}: \sys with only deduplicated prefill enabled, and without prompt assignment and no actor scaling; 3) \sys \textit{w/o AS}: \sys with no actor scaling; and 4) \sys (full version): \sys with all three core features enabled. We further compare these variants against a VERL baseline and an oracle baseline, where the oracle simulates optimal scheduling with 100\% accurate length estimates.

We visualize generation cost and execution time in Fig.~\ref{fig:eval_ablation}. Each point represents one training step; more points closer to the lower-left indicate better performance with overall lower cost and shorter time.
Fig.~\ref{fig:eval_ablation}(b) shows the cost reduction from deduplicated prefill relative to the raw version in Fig.~\ref{fig:eval_ablation}(a). While the reduction is modest in this setup, its magnitude is dataset- and algorithm-dependent and can be larger under different conditions.
Fig.~\ref{fig:eval_ablation}(c) demonstrates that prompt assignment with cut-and-migrate further reduces GPU cost by grouping prompts of similar length within the same actor. In rare cases, if cut-and-migrate is triggered late, migrations can introduce extra latency (\eg, the rightmost outlier). Overall, transmission overhead is typically overlapped with computation and does not increase end-to-end latency.
Fig.~\ref{fig:eval_ablation}(d) illustrates the speedup from actor scaling, which dynamically adjusts the number of actors to match workload demands. When additional resources are allocated for heavier steps, the shorter execution time offsets the added resources, so the total cost grows slowly or remains comparable while achieving higher throughput.

\subsection{Effectiveness of \sys}
\label{subsec:effectiveness}

To better explain how each core feature of \sys works and where its benefits come from, we run controlled experiments that enable only one feature at a time. We use the same dataset and algorithm settings as in \S\ref{subsec:ablation}.

\begin{figure}[t]
  \centering
  \includegraphics[width=\linewidth]{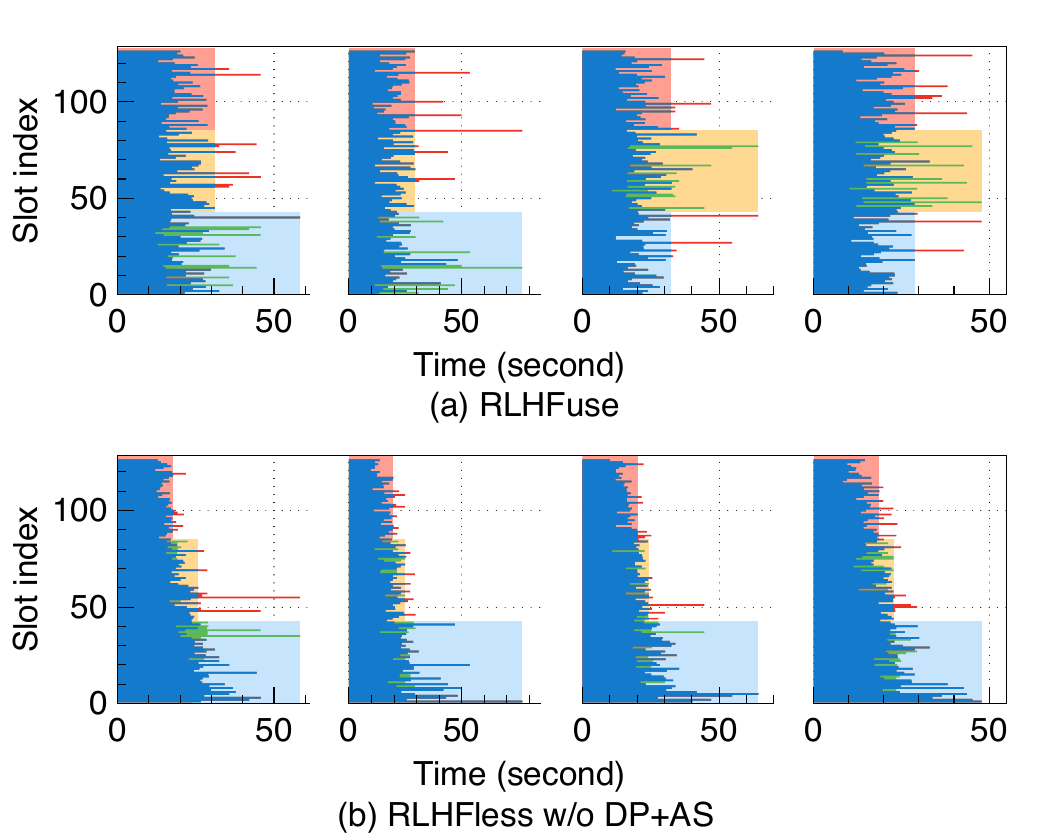}
  \vspace{-0.25in}
  \caption{Effectiveness of \sys's prompt assignment. (DP: \underline{D}eduplicated \underline{P}refill; AS: \underline{A}ctor \underline{S}caling; PA: \underline{P}rompt \underline{A}ssignment.)} 
  \vspace{-0.1in}
  \label{fig:eval_effectiveness_prompt_assignment}
\end{figure}

\noindent\textbf{Effectiveness of prompt assignment.}
\ricky{Fig.~\ref{fig:eval_effectiveness_prompt_assignment} shows the per-actor workloads and execution times of \sys compared with RLHFuse~\cite{zhong2024rlhfuseefficientrlhftraining}.}
Note that in \RLHF it is challenging to reproduce identical generated content due to hardware-level precision loss~\cite{yuan2025fp32deathchallengessolutions,he2025nondeterminism}.  
Thus, we run \sys with only prompt assignment enabled using \PPO on GSM8K, record the execution traces, and use them to simulate the raw version with all features disabled.
We report results randomly selected from four consecutive training steps.
%
In Fig.~\ref{fig:eval_effectiveness_prompt_assignment}(a), RLHFuse incurs higher GPU time because it applies a global cut-and-migrate strategy without prompt ranking, leading to uneven prompt distribution across the three actors.
By contrast, Fig.~\ref{fig:eval_effectiveness_prompt_assignment}(b) shows that enabling prompt assignment groups prompts with similar expected lengths within the same actor, allowing actors 2 and 3 with shorter workloads to finish earlier and be released sooner, further reducing the costs.
As a result, \sys's prompt assignment achieves 12.8\% end-to-end cost reduction compared to RLHFuse's global cut-and-migrate without ranking.
%
%
The response migration in \sys adds negligible overhead since only limited outlier prompts are misassigned to the wrong actor, which will be cut and migrated to unfinished neighbor actors with available slots.

\begin{figure}[t]
  \centering
  \includegraphics[width=0.9\linewidth]{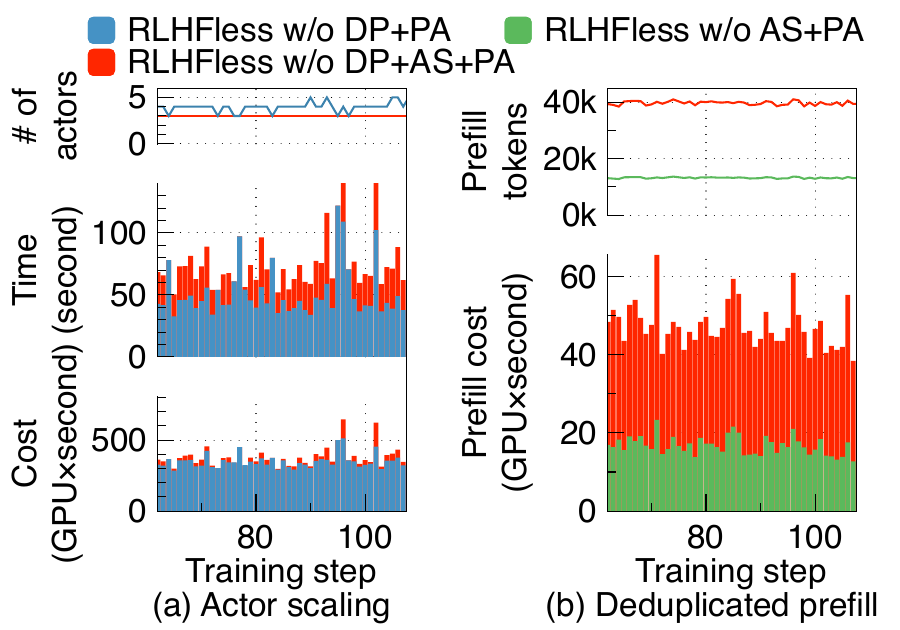}
  \vspace{-0.1in}
  \caption{Effectiveness of \sys's (a) actor scaling and (b) deduplicated prefill. (DP: \underline{D}eduplicated \underline{P}refill; AS: \underline{A}ctor \underline{S}caling; PA: \underline{P}rompt \underline{A}ssignment.)} 
  \vspace{-0.2in}
  \label{fig:eval_effectiveness_actor_scaling_deduplicated_prefill}
\end{figure}

\noindent\textbf{Effectiveness of actor scaling.}
We use the same setup described in \S\ref{subsec:ablation} and compare the raw version of \sys with the same system with only actor scaling enabled for a fair comparison.
Fig.~\ref{fig:eval_effectiveness_actor_scaling_deduplicated_prefill}(a) shows how the number of actors adapts when scaling is enabled.
With scaling, \sys often finds a sweet spot where adding actors shortens the step time enough that the total cost, computed as \# of GPUs $\times$ execution time after speeding up, is lower than using a fixed actor count.
The results show that \sys can effectively estimate the next step’s workload by referring to historical results, evaluate candidate actor counts by simulating their cost and time, and choose a setting that speeds up training without increasing cost in most cases.

\noindent\textbf{Effectiveness of deduplicated prefill.}
To evaluate deduplicated prefill in \sys, we use the same setup as \S\ref{subsec:ablation} and compare two variants: the raw \sys with no core features enabled and \sys with only deduplicated prefill enabled.
We measure, for each training step, the total number of tokens processed and the total cost of the prefill stage during \RLHF generation.
Fig.~\ref{fig:eval_effectiveness_actor_scaling_deduplicated_prefill}(b) shows that \sys reduces the amount of prefill computation by eliminating repeated KV cache calculation.
The magnitude of cost reduction depends on the \RLHF configuration. For example, in Fig.~\ref{fig:eval_effectiveness_actor_scaling_deduplicated_prefill}(b), we run \GRPO with three responses per prompt; computing the prefill once per prompt theoretically yields a 66\% reduction in prefill cost.

\subsection{Sensitivity Analysis}
\label{subsec:sensitivity}

\begin{figure}[t]
  \centering
  \includegraphics[width=0.85\linewidth]{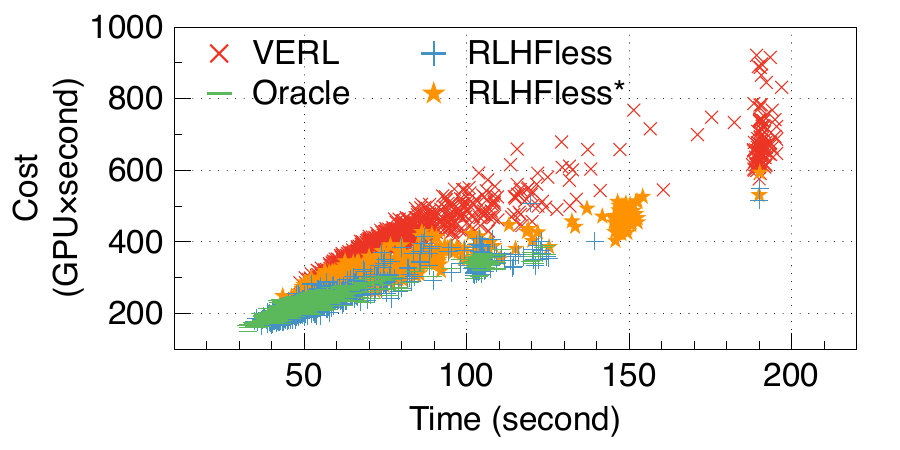}
  \vspace{-0.15in}
  \caption{Sensitivity analysis of \sys's response length prediction.} 
  \vspace{-0.25in}
  \label{fig:eval_prediction_sensitivity}
\end{figure}

\begin{figure}[t]
  \centering
  \includegraphics[width=0.8\linewidth]{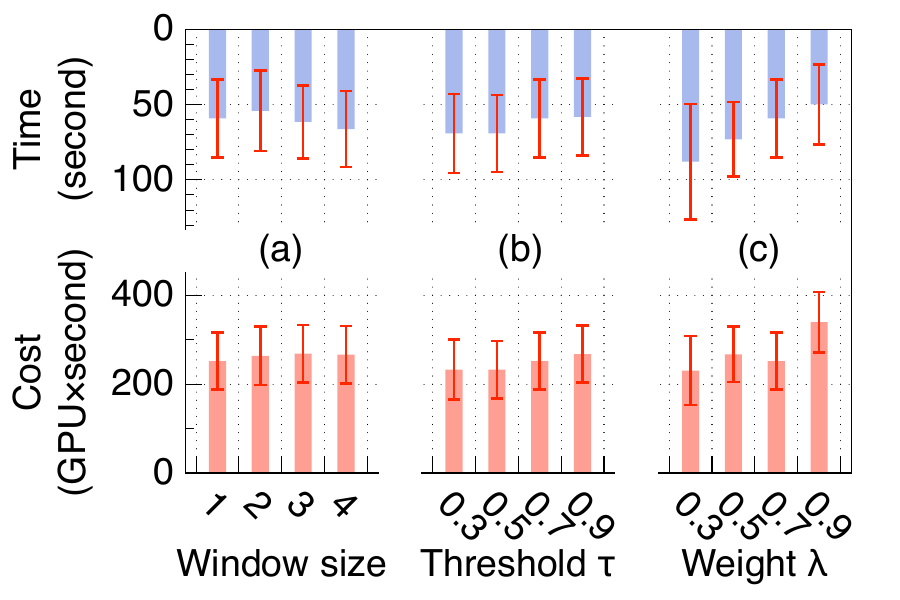}
  \vspace{-0.15in}
  \caption{\sys hyperparameter sensitivity: (a) \EWMA window size for length prediction, (b) cut-and-migrate threshold $\tau$, and (c) actor scaling weight $\lambda$.} 
  \label{fig:eval_hyperparam_sensitivity}
\end{figure}

To test robustness, justify our choices, and show how prediction accuracy and hyperparameter settings affect performance, we repeat the setup mentioned in \S\ref{subsec:ablation} and run experiments to analyze the sensitivity of each hyperparameter. We conduct the average per-step cost and time for each setting, shown in Figs.~\ref{fig:eval_prediction_sensitivity} and \ref{fig:eval_hyperparam_sensitivity}.

\noindent\textbf{Sensitivity to response length prediction accuracy.}
In \sys, both actor scaling and prompt assignment depend on response length estimates.
To test robustness against imperfect estimates, we replace our historical, training-step-based estimator with lower-accuracy predictors from prior work~\cite{qiu2024efficientinteractivellmserving,fu2024efficientllmschedulinglearning}, and evaluate the resulting performance.
We integrate a model-based predictor into \sys, denoted as {$\text{\sys}^*$}. Following prior works in \LLM serving~\cite{qiu2024efficientinteractivellmserving,fu2024efficientllmschedulinglearning}, this predictor is attached to the policy model under training. It must be pre-trained on the target \RLHF dataset, which takes approximately 11 GPU-hours, and achieves only about 44\% accuracy.
In principle, higher prediction accuracy should yield better decisions for actor scaling and prompt assignment. Fig.~\ref{fig:eval_prediction_sensitivity} shows that {$\text{\sys}^*$} still achieves 1.22$\times$ speedup and reduces \RLHF generation cost by 28.9\% over the VERL baseline, despite its lower accuracy.
This indicates that the predictor in \sys is modular and replaceable, and that \sys tolerates inaccurate estimates. The robustness comes from three factors: (1) \sys's prompt assignment relies on the ranking of expected lengths rather than exact values, (2) the cut-and-migrate backoff corrects harmful misassignments at runtime, and (3) actor scaling uses profiled \TPOT latency curves, which dampens the effect of individual prediction errors.

\noindent\textbf{Sensitivity to \EWMA window size for length prediction.}
The window size is the number of past epochs to predict next-epoch lengths. We run fifteen epochs and vary the window from 1 to 4.  
Fig.~\ref{fig:eval_hyperparam_sensitivity}(a) shows only marginal gains from larger windows. With few training epochs and often monotonic length trends, older data becomes stale and does not improve accuracy.  
We therefore set the window to 1 (use the latest epoch only) for simplicity.

\noindent\textbf{Sensitivity to cut-and-migrate threshold $\tau$.}
The threshold $\tau$ controls when to trigger the cut-and-migrate process, which terminates unfinished responses in one actor and moves them to other available actors.
We run three epochs and vary $\tau \in [0.3, 0.9]$.  
As shown in Fig.~\ref{fig:eval_hyperparam_sensitivity}(b), smaller $\tau$ triggers earlier cuts, which reduces cost by eliminating idle time in a finer granularity, but increases per-step time due to more migrations and state transfers.  
Additionally, costs do not decrease further when $\tau \le 0.5$. Even if cut-and-migrate is triggered earlier, it can proceed only when other actors have available slots, so early cuts often end up waiting and yield limited savings.
Thus, we choose $\tau=0.7$ to balance time and cost.

\noindent\textbf{Sensitivity to actor-scaling weight $\lambda$.}
The actor-scaling strategy uses a weight $\lambda$ to balance cost versus speed.  
We run three epochs and vary $\lambda \in [0.3, 0.9]$. A higher $\lambda$ prioritizes speed; a lower $\lambda$ prioritizes cost.  
Although $\lambda=0.5$ may seem like the most balanced setting, Fig.~\ref{fig:eval_hyperparam_sensitivity}(c) shows that $\lambda=0.7$ yields the best trade-off in our experiments, achieving the second-lowest cost and second-shortest step time.  
This is because the min-max normalization used in the objective does not fully preserve the original distribution of cost and time. Even after scaling to $[0,1]$, their influence on the final score may still be uneven.  
While these hyperparameters may not generalize across all workloads, the observed trends are consistent and provide a practical starting point for tuning.


\subsection{Scalability Analysis}
\label{subsec:scalability}

\begin{figure}[t]
  \centering
  \includegraphics[width=0.9\linewidth]{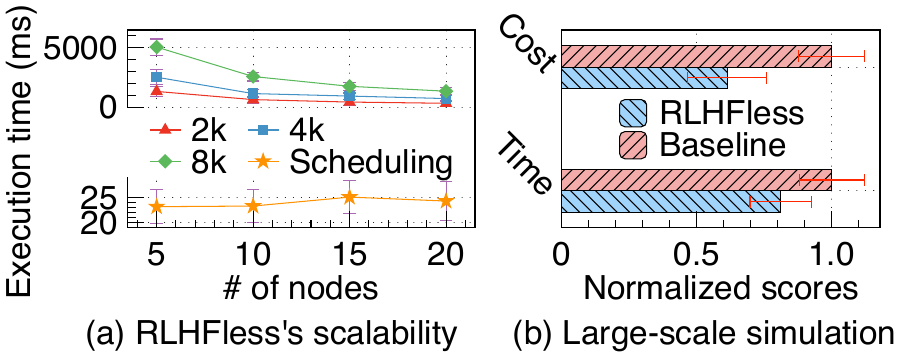}
  \vspace{-0.15in}
 \caption{\sys scalability analysis. (a) Execution time for response generation with lengths from 2k to 8k tokens, along with scheduling overhead, under different cluster sizes. (b) Average normalized performance improvement from large-scale simulation experiments.}
  \vspace{-0.1in}
  \label{fig:eval_scalability}
\end{figure}

To evaluate \sys at a larger scale under a limited budget, we implement a simulator for \PPO generation on GSM8K using Vidur~\cite{agrawal2024vidurlargescalesimulationframework}, a widely used \LLM serving simulator, together with \RLHF training traces from VERL~\cite{Sheng2025hybridflow}.
We simulate a cluster with up to 20 nodes, each equipped with 8 NVIDIA H100 GPUs connected via NVLink.
\noindent\textbf{\sys's scalability.}
We use the Llama2-70B model with a generation batch size of 512 and scale the number of nodes, allocating two GPUs per actor.
Fig.~\ref{fig:eval_scalability}(a) shows that per-step generation time decreases as more resources are added under a fixed workload, indicating strong scaling.
Meanwhile, the scheduling overhead of \sys remains stable and within 30~ms across all configurations.


\noindent\textbf{Large-scale simulation performance.}
We then calculate the average returns of \sys across the simulation settings used previously, and report the normalized results in Fig.~\ref{fig:eval_scalability}(b). As a result, \sys achieves 1.23$\times$ speedup and 38.7\% cost reduction compared to the simulated \RLHF generation baseline.

\subsection{Latency Breakdown}
\label{subsec:latency_breakdown}

\begin{figure}[t]
  \centering
  \includegraphics[width=0.8\linewidth]{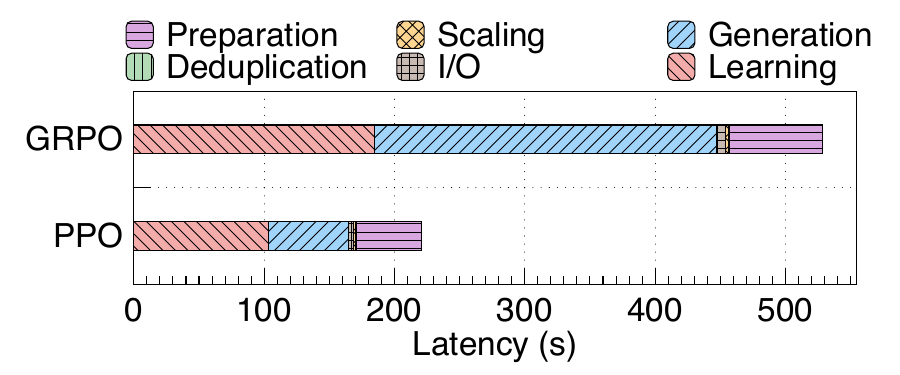}
  \vspace{-0.15in}
  \caption{\sys's latency breakdown in a training step, including the preparation, generation, and learning phases, as well as deduplication latency, scaling latency (response-length prediction and cost/time estimation), and I/O latency.} 
  \vspace{-0.2in}
  \label{fig:eval_latency_breakdown}
\end{figure}

We evaluate the latency breakdown and overhead of \sys's core components within a single \RLHF training step.  
Fig.~\ref{fig:eval_latency_breakdown} presents the breakdown for two representative \RLHF algorithms: \PPO and \GRPO.  
The overhead introduced by \sys's core features, including prompt deduplication, response length prediction, and cost/time estimation, is negligible relative to the total step time.  
This is because all core components in \sys are model-free and lightweight, adding minimal computational burden to the training pipeline.

\section{Related Work}

\noindent\textbf{\LLM alignment methods.} 
After \SFT, \LLM alignment methods can be broadly divided into \RL-based and non-\RL approaches~\cite{wang2024comprehensivesurveyllmalignment}.
Common \RL-based methods include \PPO~\cite{schulman2017proximalpolicyoptimizationalgorithms,zheng2023rlhfppo}, \GRPO~\cite{shao2024grpo}, ReMax~\cite{li2024remaxsimpleeffectiveefficient}, REINFORCE~\cite{ahmadian2024basicsrevisitingreinforcestyle}, and REINFORCE++~\cite{hu2025reinforce++}.
Non-\RL methods such as \DPO~\cite{rafailov2024directpreferenceoptimizationlanguage} remove the RL loop by reformulating preference learning as supervised optimization, but have been shown to underperform \RL-based methods on complex reasoning and coding tasks~\cite{xu2024dposuperiorppollm}.
In this paper, we identify unique patterns of \RL-based methods and design corresponding optimizations to improve the resource efficiency and accelerate the sampling process.


\noindent\textbf{\RL and \LLM alignment frameworks.} 
Traditional \RL frameworks~\cite{liang2018rllibabstractionsdistributedreinforcement,hoffman2020acme,huang2022cleanrl,MSRL} target small-scale models and do not scale to \LLM alignment due to model size and communication overhead.
Recent \LLM alignment frameworks~\cite{Collosal-Chat,yao2023deepspeedchateasyfastaffordable,shen2024nemoalignerscalabletoolkitefficient,Sheng2025hybridflow,hu2024openrlhfeasytousescalablehighperformance,zhong2024rlhfuseefficientrlhftraining,slime_github,fu2025areallargescaleasynchronousreinforcement,li2026a3poacceleratingasynchronousllm,griggs2025skrylv01} typically combine a high-throughput inference engine~\cite{kwon2023vllm,TensorRT-LLM,zheng2024sglangefficientexecutionstructured} with a distributed training backend~\cite{shoeybi2020megatronlmtrainingmultibillionparameter,deepspeed,zhao2023pytorchfsdpexperiencesscaling}.
To mitigate idle time in staged \RLHF workflows, some systems adopt asynchronous and off-policy execution~\cite{li2026a3poacceleratingasynchronousllm,fu2025areallargescaleasynchronousreinforcement,slime_github,griggs2025skrylv01}, although prior work shows that stale data can degrade training quality~\cite{zheng2025prosperitycollapsefaroffpolicy,noukhovitch2025asynchronousrlhffasterefficient}.
Meanwhile, some systems optimize synchronous training through model placement and parallelism~\cite{yao2023deepspeedchateasyfastaffordable,shen2024nemoalignerscalabletoolkitefficient,hu2024openrlhfeasytousescalablehighperformance,Sheng2025hybridflow}. 
All of these frameworks rely on serverful infrastructure and cannot adapt resource allocation at fine granularity across training steps.
RLHFuse (FlexFusion)~\cite{zhong2024rlhfuseefficientrlhftraining} offloads long-tail responses to dedicated GPUs, but its global cut-and-migrate strategy may incur extra overhead without workload-aware prompt assignment, as discussed in \S\ref{subsec:effectiveness}.
In contrast, \sys introduces serverless execution with deduplicated prefill, actor scaling, and prompt assignment, which are orthogonal to existing synchronous frameworks.

\noindent\textbf{Serverless computing for \LLM.}
Serverless computing has gained increasing attention in \LLM applications due to its flexibility, automatic scaling, and pay-as-you-go pricing model, especially in the context of \LLM serving~\cite{fu2024serverlessllm,liu2025moe,yu2025fmoe,tao2024ENOVA}. 
ServerlessLLM~\cite{fu2024serverlessllm} supports low-latency \LLM serving by leveraging the large memory and storage capacities available on modern GPU servers. 
ENOVA~\cite{tao2024ENOVA} introduces a configuration recommendation module for automatic deployment across various GPU clusters, along with a performance monitoring module that enables auto-scaling. 
However, all of these serverless systems are focused solely on \LLM serving and they are not designed for \RLHF workloads.

\section{Conclusion}
This paper presents \sys, the first serverless framework for RLHF training.
By targeting key inefficiencies in serverful RLHF systems---including idle time across dependent training phases, dynamic response lengths, and redundant KV cache computation---\sys leverages serverless execution for fine-grained resource adaptation.
Through deduplicated prefill, prompt assignment, and cost-aware actor scaling, \sys reduces resource wastage both across the RLHF loop and within individual components.
Experiments show that \sys achieves up to $1.35\times$ speedup with $44.8\%$ cost reduction on physical clusters, and an average $1.23\times$ speedup with $38.7\%$ cost reduction on a simulated large-scale GPU cluster, demonstrating the effectiveness of serverless RLHF training.

\bibliographystyle{ACM-Reference-Format}
\bibliography{reference}

\end{document}